\documentclass[letterpaper, 10 pt, conference]{ieeeconf}  

\IEEEoverridecommandlockouts                              

\usepackage{times}
\usepackage{amsmath}
\usepackage{amssymb}
\usepackage{amsthm}
\usepackage{graphics}
\usepackage{epsfig}
\usepackage{multirow}
\usepackage{nicefrac}
\usepackage{cases}          
\usepackage{subcaption}
\usepackage{mathtools}
\usepackage{xcolor}
\usepackage{algorithm}
\usepackage{algorithmicx}
\usepackage{algpseudocode}
\usepackage{balance}
\usepackage{gensymb}

\newcommand{\tn}[1]{\textnormal{#1}}
\newcommand{\tb}[1]{\textbf{#1}}

\newcommand{\mat}[0]{\begin{bmatrix}}
\newcommand{\mate}[0]{\end{bmatrix}}

\newcommand{\vf}{\mathbf{f}}
\newcommand{\vg}{\mathbf{g}}
\newcommand{\vh}{\mathbf{h}}

\newcommand{\vp}{\mathbf{p}}

\newcommand{\vu}{\mathbf{u}}
\newcommand{\vv}{\mathbf{v}}

\newcommand{\vx}{\mathbf{x}}

\newcommand{\vz}{\mathbf{z}}

\newcommand{\cD}{\mathcal{D}}

\newcommand{\cH}{\mathcal{H}}
\newcommand{\cI}{\mathcal{I}}

\newcommand{\cS}{\mathcal{S}}
\newcommand{\cT}{\mathcal{T}}

\newcommand{\cW}{\mathcal{W}}

\newcommand{\R}{\mathbb{R}}
\newcommand{\N}{\mathbb{N}}

\newcommand{\X}{\mathbb{X}}
\newcommand{\U}{\mathbb{U}}

\newcommand\norm[1]{\left\|#1\right\|}              

\makeatletter
\let\NAT@parse\undefined
\makeatother

\usepackage{hyperref}
\hypersetup{citebordercolor=green}
\hypersetup{linkbordercolor=green}
\hypersetup{urlbordercolor=green}
\hypersetup{hidelinks}
\usepackage{array}
\newcolumntype{L}[1]{>{\raggedright}m{#1}}
\newcolumntype{C}[1]{>{\centering}m{#1}}
\newcolumntype{R}[1]{>{\raggedleft}m{#1}}

\newcommand{\rebuttal}[1]{{\color{black}#1}}

\graphicspath{{images/}}

\title{\LARGE \bf
Learning Interaction-Aware Trajectory Predictions for Decentralized Multi-Robot Motion Planning in Dynamic Environments
}
\author{Hai Zhu$^*$, Francisco Martinez Claramunt$^*$, Bruno Brito and Javier Alonso-Mora%
\thanks{$^*$The authors contributed equally. } 
\thanks{This work is supported in part by the U.S. Office of Naval Research Global (ONRG) NICOP-grant N62909-19-1-2027, the Amsterdam Institute for Advanced Metropolitan Solutions (AMS) and the Netherlands Organization for Scientific Research (NWO) domain Applied Sciences (Veni 15916). The authors are with the Department of Cognitive Robotics, Delft University of Technology, 2628
CD Delft, The Netherlands \{\tt\small h.zhu; bruno.debrito; j.alonsomora\}@tudelft.nl. fmartinezclaramunt@outlook.com} %
} 
\begin{document}
\maketitle              


\begin{abstract} 

This paper presents a data-driven decentralized trajectory optimization approach for multi-robot motion planning in dynamic environments. When navigating in a shared space, each robot needs accurate motion predictions of neighboring robots to achieve predictive collision avoidance. These motion predictions can be obtained among robots by sharing their future planned trajectories with each other via communication. However, such communication may not be available nor reliable in practice. In this paper, we introduce a novel trajectory prediction model based on recurrent neural networks (RNN) that can learn multi-robot motion behaviors from demonstrated trajectories generated using a centralized sequential planner. The learned model can run efficiently online for each robot and provide interaction-aware trajectory predictions of its neighbors based on observations of their history states. We then incorporate the trajectory prediction model into a decentralized model predictive control (MPC) framework for multi-robot collision avoidance. Simulation results show that our decentralized approach can achieve a comparable level of performance to a centralized planner while being communication-free and scalable to a large number of robots. We also validate our approach with a team of quadrotors in real-world experiments. 

\end{abstract}

\section{Introduction}\label{sec:intro}
Autonomous navigation of a team of robots in dynamic environments is important when deploying them in various applications such as coverage and inspection \cite{breitenmoser2016combining}, search and rescue \cite{baxter2007multi}, formation flying \cite{Zhu2019ICRA} and multi-view videography \cite{Nageli2017}. In these scenarios, the robots navigate in a shared space that may also have moving obstacles. To achieve predictive collision avoidance and ensure safety, each robot needs to know the future motion predictions of other robots in the environment. These motion predictions can be obtained among robots by sharing their future planned trajectories with each other via communication \cite{Zhu2019RAL}. However, such communication may not be available nor reliable in practice. Alternatively, some approaches \cite{Kamel2017} employ a constant velocity model to predict other robots' trajectories. Even though communication among robots is not required, the planned robot motions may not be safe, particularly in crowded dynamic environments \cite{Zhu2019RAL}. 

\begin{figure}[t]
	\centering
	\setlength{\textfloatsep}{0.7\baselineskip plus 0.2\baselineskip minus 0.5\baselineskip}
	\includegraphics[width=0.86\linewidth]{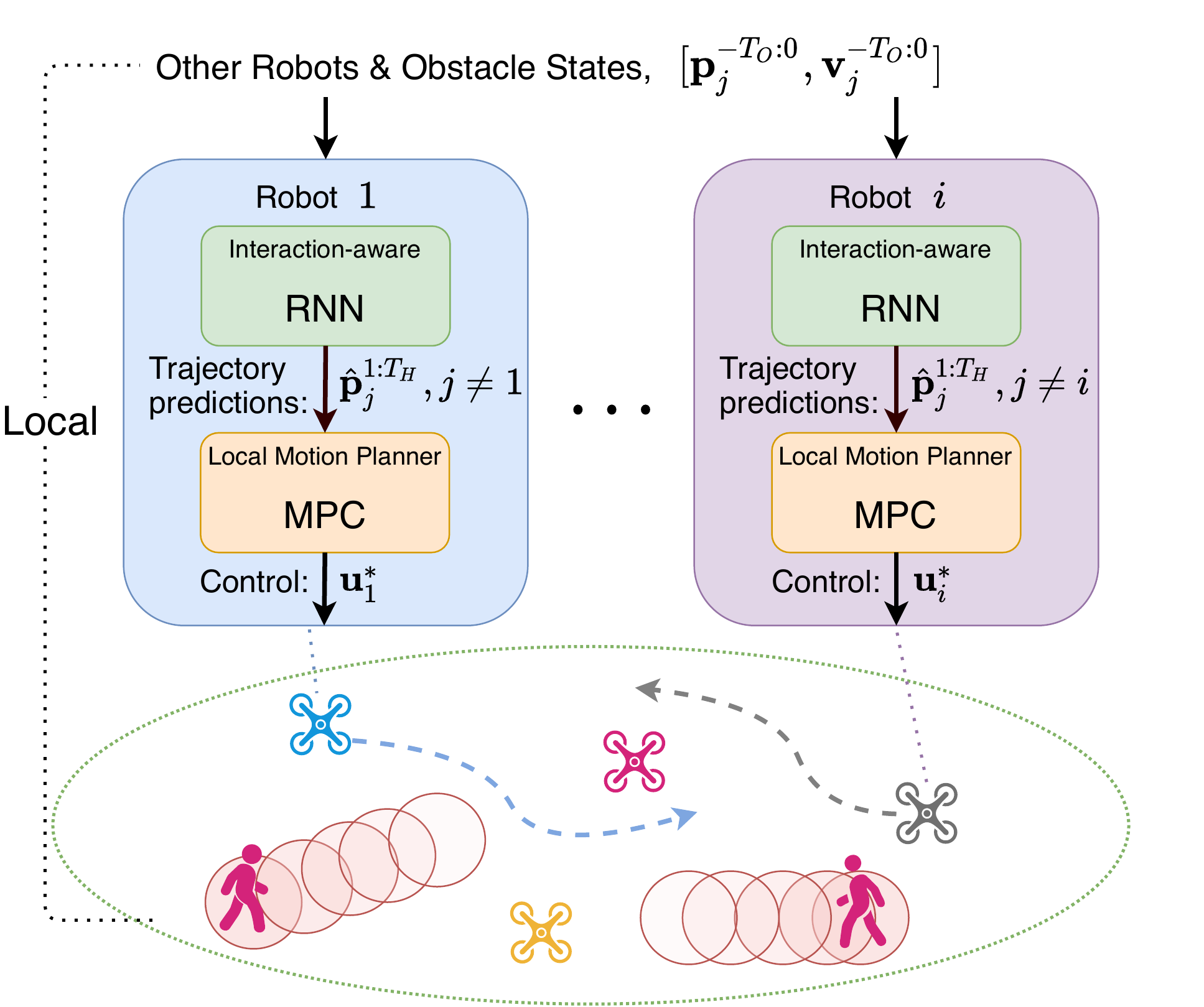}
	\caption{The proposed decentralized communication-free motion planner that relies on a RNN model for interaction-aware trajectory prediction and a MPC for local motion planning.}
	\label{fig:overview}
\end{figure}

In this paper, we propose an interaction- and obstacle-aware trajectory prediction model and combine it with the model predictive control (MPC) framework to achieve decentralized multi-robot motion planning in dynamic environments. Fig. \ref{fig:overview} gives an overview of the proposed method. In particular, we first generate a demonstration dataset consisting of robot trajectories using a multi-robot collision avoidance simulator \cite{Zhu2019RAL}. It utilizes a centralized sequential MPC for local motion planning in which inter-robot communication is employed. Next, we formulate the robot trajectory prediction problem as a sequence modeling task and hence design a model based on recurrent neural networks (RNN).
By training the model using the generated dataset, it learns to imitate the centralized sequential MPC and thus can predict the planning behaviors of the robots. 
Finally by combining the trajectory prediction model with the MPC framework, multi-robot local motion planning is achieved in a decentralized manner. 

The main contributions of this work are:
\begin{itemize}
    \item A RNN-based interaction- and obstacle-aware model that is able to provide robot trajectory predictions in a multi-robot scenario.
    \item Incorporation of the trajectory prediction model with MPC to achieve decentralized multi-robot local motion planning in dynamic environments. 
\end{itemize}

We show that our designed model can make accurate trajectory predictions, thanks to which the proposed decentralized multi-robot motion planner can achieve a comparable level of performance to the centralized planner while being communication-free. We also validate our approach with a team of quadrotors in real-world experiments.

\section{Related Work}\label{sec:relatedWork}

\subsection{Multi-Robot Collision Avoidance}
We focus our work on online local motion planning for multi-robot systems (also referred as multi-robot collision avoidance), which has been actively studied over the past years. 
Traditional reactive controller-level approaches include the optimal reciprocal collision avoidance (ORCA) method \cite{vandenBerg2011} that builds on the concept of reciprocal velocity obstacles (RVO) \cite{vandenBerg2008}, artificial potential field (APF) based method \cite{yongjie2009collision}, buffered Voronoi cell (BVC) approach \cite{Zhou2017,Zhu2019MRS}, and control barrier functions (CBF) \cite{Wang2017}. While these \rebuttal{reactive} methods are computationally efficient, the robot dynamics are not fully modeled and the robot motion is typically limited by only planning one time step ahead. Recently, there have emerged new learning-based methods for multi-robot collision avoidance, such as deep imitation learning \cite{Shi2020,Riviere2020} and those that are reinforcement learning (RL) based \cite{Chen2017a,semnani2020multi}. 
\rebuttal{
RL-based methods can learn policies that have a long-term cumulative reward for the robots and thus are considered to be non-myopic \cite{Chen2017a}. However, they are generally not able to handle hard state constraints, such as collision avoidance constraints.} 
These issues can be overcome by using the model predictive control (MPC) framework for collision-free trajectory generation in which an optimization problem is solved for each robot in a receding horizon manner. In this paper, we study the multi-robot MPC-based collision avoidance problem. 

For each robot to solve a local trajectory optimization problem, it needs to know the future trajectories of other robots. One approach is to let each robot communicate its planned trajectory with every other robot in the team. Hence, robots can then update their own trajectories to be collision free with other robots' trajectory plans, as in these distributed MPC works \cite{Zhu2019RAL,Luis2020}. While these methods can achieve safe collision avoidance, the communication burden across the team is large and may not be available nor reliable in practice \cite{Serra2020}. Another approach is to let each robot predict other robots' future motions based on its own observations. For instance, \cite{Kamel2017} employs a constant velocity model when predicting other robots' future trajectories. In that case, communication among robots is not required. However, such a prediction can be inaccurate and may lead to unsafe trajectory planning \cite{Zhu2019RAL}. In this paper, we will develop an interaction- and obstacle-aware model for the trajectory prediction taking into account surrounding information of the robot to model the interaction and environment constraints. By incorporating the model with the MPC framework, we can achieve safe and communication-free decentralized collision avoidance for multiple robots in dynamic environments.

\subsection{Motion Prediction}
\rebuttal{
    Our proposed approach decouples motion prediction and trajectory planning to achieve decentralized and communica- tion-free collision avoidance. Such a decoupling is also seen in \cite{Schmerling2018,Fridovich-Keil2020}, where the motion prediction of humans are used to plan a safe trajectory for the ego robot.}
Motion prediction for decision-making agents has drawn significant research efforts over the past years, with most works focusing on human trajectory prediction \cite{Rudenko2020}. Early works on motion prediction are typically model-based such as the renowned social force-based method \cite{Helbing1995} which models pedestrian behaviors through the use of attractive and repulsive potentials. The model is later generalized and adapted to modeling traffic car behaviors \cite{helbing1998generalized}. While these methods are computationally efficient, the prediction accuracy is quite low. There have also been several notable attempts to utilize game theory to model interacting decision-making agents and predict their future trajectories \cite{Turnwald2016, Oyler2016}, in which the agents are assumed to play a non-cooperative game and their trajectory predictions can be obtained from computing the Nash equilibria of the game. 
While interaction-aware trajectory predictions can be obtained, these methods are limited to specific road scenarios and cannot be directly applied to general multi-robot systems. 

The class of approaches that have achieved state-of-the-art performance in trajectory prediction problems are the learning-based methods. Some of these include inverse reinforcement learning (IRL) \cite{Kuderer2015}, recurrent neural networks (RNN) \cite{Alahi2016,Brito2020}, variational autoencoders \cite{Lee2017}, generative adversarial networks (GAN) \cite{Gupta2018} \rebuttal{that provide predicted human trajectories in two-dimensional (2D) environments, Gaussian mixture regression (GMR) \cite{mainprice2013human} and Gaussian process regression (GPR) \cite{park2019planner} that can predict human actions in 3D workspaces.}
\rebuttal{
    Our approach of predicting the trajectories of other robots is based on previous works on human motion prediction since both can be formulated as a sequence modelling problem. In particular, our prediction model is based on RNN, inspired by the works in \cite{Pfeiffer2018} for interaction-aware pedestrian motion prediction in which static obstacles are considered and represented using a grid map. We adapt the model to predict robots trajectories in multi-robot scenarios with moving obstacles described by their positions and velocities, and further apply the model to decentralized multi-robot motion planning by incorporating it within MPC. 
}

\section{Preliminaries}\label{sec:preliminary}
Throughout this paper vectors are denoted in bold lowercase letters, $\vx$, matrices in plain uppercase, $M$, and sets in calligraphic, $\cS$. $\norm{\vx}$ denotes the Euclidean norm of $\vx$ and $\norm{\vx}_{Q}^{\rebuttal{2}} = \vx^TQ\vx$ denotes the weighted squared norm.

\subsection{Robot and Obstacle Model}\label{subsec:robot_obs_model}
Following \cite{Zhu2019RAL}, we consider a team of $n$ robots moving in a shared workspace $\cW \subseteq \R^3$, where each robot $i \in \cI = \{ 1,2,\dots, n \} \subset \N $ is modeled as an enclosing sphere with radius $r$. The robots follow the same dynamical model that is described by a discrete-time equation as follows,
\begin{equation}\label{eq:nonDyn}
    \begin{aligned}
            \vx_i^{k+1} = \vf(\vx_i^k, \vu_i^k), \quad \vx_i^0 = \vx_i(0),
    \end{aligned}
\end{equation}
where $\vx_i^k \in \X \subset \R^{n_x}$ denotes the state of the robot, typically including its position $\vp_i^k$ and velocity $\vv_i^k$, and $\vu_i^k \in \U \subset \R^{n_u}$ the control inputs at time $k$. Without loss of generality, $k=0$ indicates the current time. $\X$ and $\U$ are the admissible state space and control space, respectively. $\vx_i(0)$ is the current state of robot $i$. 
In addition, moving obstacles for example pedestrians in the environment are considered. For each obstacle $o \in \cI_o = \{1,2,\dots,n_o \} \subset \N $ at position $\vp_o \in \R^3$, we model it as an upright non-rotating enclosing \emph{ellipsoid} centered at $\vp_o$ with semi-principal axes $(a, b, c)$. 

\rebuttal{
In this paper, we assume that each robot can observe the states (positions and velocities) of all other robots and moving obstacles and keep their history information.}

\subsection{Multi-Robot Collision Avoidance}\label{subsec:mrca}
Multi-robot local motion planning is considered in this paper, in which the goal is to achieve real-time collision-free navigation for multiple robots. Each robot has a given goal location $\vg_i$, which generally comes from some high-level path planner \cite{Honig2018} or is specified by the user. Any pair of robots $i$ and $j$ from the group are mutually collision-free if $\norm{\vp_i^k - \vp_j^k} \geq 2r, \forall i\neq j \in \cI, k = 0, 1, \dots $. Regarding robot-obstacle collision avoidance, we approximate the obstacle with an enlarged ellipsoid and check if the robot's position is inside it. Hence, the robot $i$ is collision-free with the obstacle $o$ at time step $k$ if $\norm{\vp_i^k-\vp_o^k}_{\Omega} \geq 1$, where $\Omega = \text{diag}(1/(a+r)^2, 1/(b+r)^2, 1/(c+r)^2)$. 

The objective is to compute a local motion $\vu_i^k$ for each robot in the group, that respects its dynamic constraints, makes progress towards its goal location $\vg_i$ and is collision-free with other robots in the group as well as moving obstacles within a planning time horizon.

\subsection{Model Predictive Control}\label{subsec:mpc}
The multi-robot collision avoidance problem can be solved using model predictive control by formulating a receding horizon constrained optimization problem. For each robot $i \in \cI$, a discrete-time constrained optimization formulation with $N$ time steps and planning horizon $N\Delta t$, where $\Delta t$ is the sampling time, is derived as follows,
\begin{subequations}\label{eq:dmpc}
	\begin{alignat}{2}
		\min_{\substack{\vx_i^{0:N}, \vu_i^{0:N-1}, \\ s^{0:N}}}  ~~        
		& \sum_{k=0}^{N-1} J_i^k(\vx_i^k, \vu_i^k, s^k) + J_i^N(\vx_i^N, \vg_i, s^N) \\
		\text{s.t.}	~~	        & \vx_i^0 = \vx_i(0), \\
		& \vx_i^{k} = \vf(\vx_i^{k-1}, \vu_i^{k-1}), \\
		& \norm{\vp_i^k - \vp_j^k} \geq 2r - s^{k}, \label{subeq:inter_robot_ca}\\ 
		& \norm{\vp_i^k - \vp_o^k}_{\Omega} \geq 1 - s^{k},\label{subeq:robot_obs_ca}\\ 
		& s^{k} \geq 0, \vu_i^{k-1} \in \U, \vx_i^k \in \X,\\
		&\forall j \neq i\in\cI; \, \forall o\in\cI_o; \, \forall k\in \{1,\dots,N\},
	\end{alignat}
\end{subequations}
where $J_i^k(\vx_i^k, \vu_i^k, s^k)$ and $J_i^N(\vx_i^N, \vg_i, s^N)$ are the stage and terminal costs, and $s$ is the slack variable. At each time step, each robot in the team solves online the constrained optimization problem (\ref{eq:dmpc}) and then executes the first step control inputs, in a receding horizon fashion. 

Note that for each robot to solve the optimization problem (\ref{eq:dmpc}), it has to know the future trajectories of other robots and moving obstacles, as shown in Eq. (\ref{subeq:inter_robot_ca}) and Eq. (\ref{subeq:robot_obs_ca}). For moving obstacles (pedestrians), we assume their motions follow a constant velocity model (CVM) in the short planning horizon and predict their future trajectories accordingly.
\rebuttal{ 
This assumption is reasonable since CVM can achieve state-of-the-art performance when used for pedestrian motion prediction \cite{Scholler2020}.}
For robots' future trajectories, denote by $\cT_{i}^0 = \{ \vp_i^{1:N} \}$ the robot $i$'s current planned trajectory. Further denote by $\hat{\cT}_{i,j}^0 = \{\vp_j^{1:N}\}$ the trajectory of robot $j\in\cI, j\neq i$ that robot $i$ assumes and uses in solving the problem (\ref{eq:dmpc}), where the hat $\hat{\cdot}$ indicates that it is robot $i$'s estimation of the other robot's trajectory. 

Typically, there are two ways for robot $i$ to obtain the future trajectory of the other robot $j$. The first way is via communication: all robots in the team communicate their planned trajectories to each other at each time step. It can be implemented using a centralized sequential planning framework as in \cite{Zhu2019RAL}, that is, $\hat{\cT}_{i,j}^0 = \cT_j^0$. Although this method guarantees collision avoidance by construction, it does not scale well with a large number of robots. Moreover, communication is not always available and reliable in practice. 

The other way is without communication. Hence, robot $i$ has to predict another robot $j$'s future trajectory based on its observation of the environment:
\begin{equation}\label{eq:prediction_general}
	\hat{\cT}_{i,j}^0 = \tn{\tb{prediction}}(\cH_i^0),
\end{equation}
where $\cH_i^0$ is the information that robot $i$ can acquire until current time from its observation. Previous works \cite{Zhu2019RAL,Kamel2017} use a constant velocity model to perform the prediction only based on the other robot's current state, that is, $\cH_i^0 = \vx_j^0$. However, such a prediction can be inaccurate and may lead to unsafe trajectory planning \cite{Zhu2019RAL}. In this paper, we will develop an interaction- and obstacle-aware model for the trajectory prediction taking into account surrounding environment information of the robot to model the interaction and environment constraints.

\section{Approach}\label{sec:method}
In this section, we present our interaction and obstacle-aware trajectory prediction method and incorporate it with the MPC framework to achieve decentralized multi-robot collision avoidance in dynamic environments.

\subsection{Trajectory Prediction Problem Formulation}
As shown in Eq. (\ref{eq:prediction_general}), robot $i\in\cI$ needs to predict the future trajectories of other robots $j \neq i \in \cI$ to plan its safe motion. Hereafter, we refer to the robot $i$ as the ego robot and the robot $j$ as the \emph{query robot} that is indicated by the sub-script $\cdot_q$. In addition, we use the sub-script $\cdot_{-q}$ to indicate the collection of all the other robots except for the query robot. 

We aim to address the problem of finding a trajectory prediction model for the query robot $q$ that gives a sequence of its future positions $\vp_q^{1:T_H}$ in a multi-robot scenario. Here $T_H \geq N$ is the prediction horizon that should not be smaller than the local motion planning horizon. As has been shown in previous trajectory prediction works \cite{Pfeiffer2018,Brito2020}, we will instead work with sequences of velocities $\vv_q^{1:T_H}$ for prediction to avoid overfitting when based on position sequences, and numerically integrate them afterwards starting from the query robot's current position $\vp_q^0$. 

Denote by $\vv_q^{-T_O:0}$ the past sequence of velocities of the query robot within an observation time $T_O \geq 1$. 
Denote by $\vp_{-q,r}^{-T_O:0}$ and $\vv_{-q,r}^{-T_O:0}$ the past relative positions and velocities of other robots with respect to the query robot. 
Further denote by $\vp_{\cI_o,r}^{0}$ and $\vv_{\cI_o,r}^{0}$ the current relative positions and velocities of the moving obstacles $o\in\cI_o$ with respect to the query robot. By observing history states of the query robot and its surrounding other robots as well as moving obstacles, we want to find an interaction- and obstacle-aware model $\vh_{\boldsymbol\theta}$ with parameters $\boldsymbol\theta$:
\begin{equation}\label{eq:prediction_formulation}
	\vv_q^{1:T_H} = \vh_{\boldsymbol\theta}(\vv_q^{-T_O:0}, \vp_{-q,r}^{-T_O:0}, \vv_{-q,r}^{-T_O:0}, \vp_{\cI_o,r}^{0}, \vv_{\cI_o,r}^{0}),
\end{equation}
that outputs a prediction of the query robot's future states.

\subsection{Demonstration Data Generation}\label{subsec:dataset}
We use a simulation dataset to train our designed network model. The dataset is generated using demonstrations from a multi-robot collision avoidance simulator \cite{Zhu2019RAL} which employs a centralized sequential planner to solve the problem (\ref{eq:dmpc}). This involves each robot solving a MPC problem sequentially and communicates its planned trajectory to other robots to avoid. 
Note that the planner differs from the prioritized planning approach since each robot has to avoid all other robots and hence it shows cooperation among robots.

Specifically, we create a three-dimensional environment in which a team of robots and moving obstacles are simulated.
In the simulation, each robot navigates to a randomly generated goal position, which is changed dynamically to a new location after being reached. The generated robots' goal positions are ensured to be collision-free with each other and the obstacles. Moving obstacles are simulated in the environment by randomly specifying an initial position and velocity \rebuttal{(with speed between 0.5 m/s and 1.2 m/s)} to each of them and then make them move at a constant velocity. Once any obstacle moves out of the environment, a new initial position and velocity will be set to it. Moreover, we add small Gaussian noise to the velocities of the moving obstacles in simulation. We perform the simulation for $N_{\tn{sim}}$ time steps and record the positions and velocities of all robots and obstacles at each time step.
After running the simulation, for each time step $t$ and robot $q$, we retrieve its future sequence of velocities and observation of the past states of the system from the recorded data. Hence, our dataset is as follows
\begin{align}
	\cD = \{ (\cH_q^t, \vv_q^{t+1:t+T_H}) | \forall q \in \cI, \forall t \in \{1,\dots, N_{\tn{sim}-T_H}\} \},
\end{align}
where the observation information $\cH_q^t$ is 
\begin{equation}
	\cH_i^t = \{\vv_q^{t-T_O:t}, \vp_{-q,r}^{t-T_O:t}, \vv_{-q,r}^{t-T_O:t}, \vp_{\cI_o,r}^{t}, \vv_{\cI_o,r}^{t}\}.
\end{equation}

\subsection{Interaction- and Obstacle-Aware Model}
\begin{figure*}[t]
	\centering
	\setlength{\textfloatsep}{0.1\baselineskip plus 0.0\baselineskip minus 0.4\baselineskip}
	\includegraphics[width=0.82\linewidth]{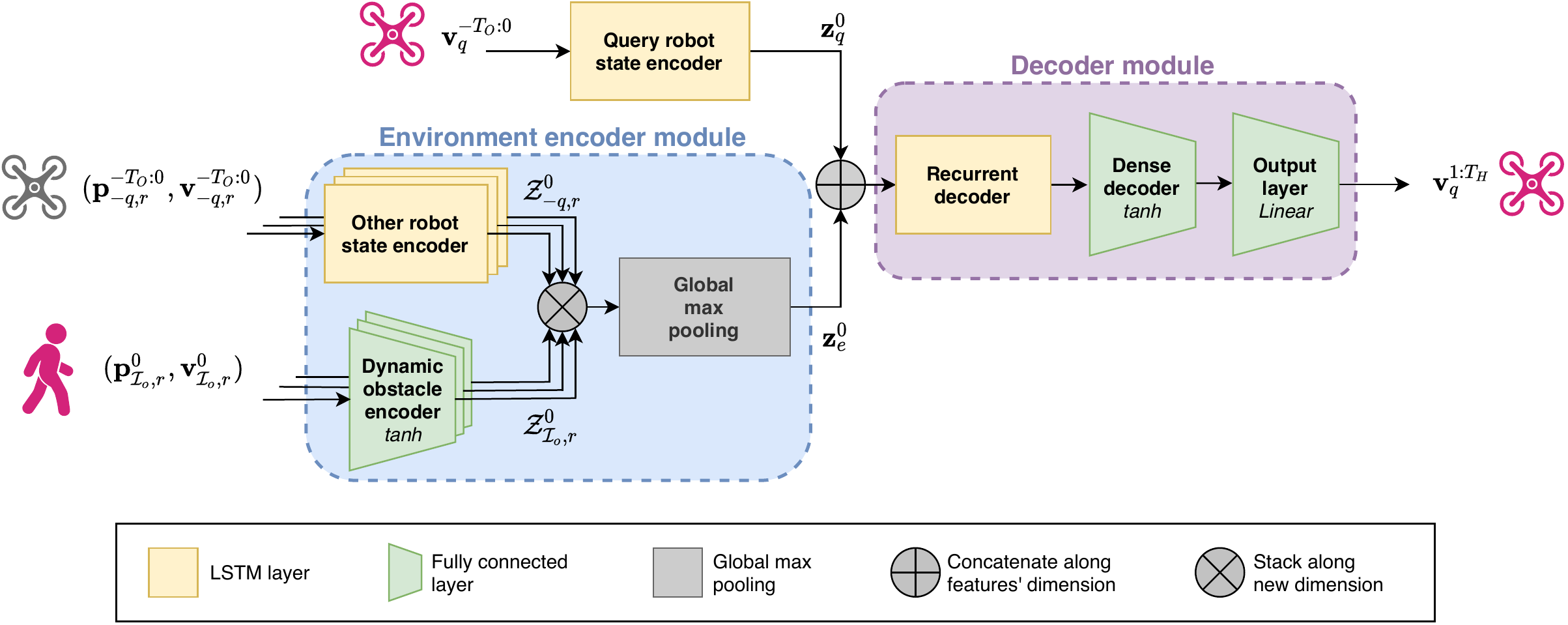}
	\caption{Network architecture of the interaction- and obstacle-aware model. Three channels of information are taken as inputs: the query robot's past velocities $\vv_q^{-T_O:0}$, past relative states of other robots $(\vp_{-q,r}^{-T_O:0}, \vv_{-q,r}^{-T_O:0})$ and current relative states of obstacles $(\vp_{\cI_o,r}^0, \vv_{\cI_o,r}^0)$. A joint representation of the inputs is created through a query robot encoder and an environment encoder. A decoder is adopted to output a sequence of velocities $\vv_q^{1:T_H}$ predicted for the query robot's future trajectory.}
	\label{fig:architecture}
\end{figure*}

We now present our recurrent neural network (RNN) model for interaction- and obstacle-aware trajectory prediction, as shown in Fig. \ref{fig:architecture}. The model first creates a joint representation of three input channels: the query robot's history state, information of other interacting robots and moving obstacles, via a query robot state encoder and an environment encoder module. Then a decoder module is adopted to output a predicted trajectory of the query robot. The recurrent layers in the model are of the LSTM type \cite{Hochreiter1997} that has been shown able to learn time dependencies over a long period of time. 
Next, we describe the three main modules of the model in detail. 


\subsubsection{Query robot state encoder} 
It consists of a recurrent layer that produces a flat encoding $\vz_q^0$ from the history velocities of the query robot $\vv_q^{-T_O:0}$.
This layer learns a dynamical model of the query robot, so that the network can leverage it to obtain better predictions.

\subsubsection{Environment encoder module}
It includes $n-1$ parallel recurrent layers with shared weights to encode the sequences of past relative positions and velocities of other robots with respect to the query robot $(\vp_{-q,r}^{-T_O:0}, \vv_{-q,r}^{-T_O:0})$ into a set $\mathcal{Z}_{-q,r}^{0}$, and $n_o$ parallel dense layers with shared weights that encode the current relative positions and velocities of moving obstacles with respect to the query robot $(\vp_{\cI_o,r}^{0}, \vv_{\cI_o,r}^{0})$ into a set $\mathcal{Z}_{\cI_o,r}^{0}$. The encodings from both of these sets, which are made to have the same length, are then stacked together and followed by a global max pooling operation executed along the new data axis. Thus, this module can capture the interaction of the query robot with a variable number of other robots and obstacles in the environment and encode it into a single flat vector $\vz_e^0$.
This framework also makes it possible to account for potentially different types of agents and obstacles by training their own set of encoders and stacking them with the rest of intermediate encodings.

\subsubsection{Decoder module}
It takes in the concatenation of $\vz_q^0$ with $\vz_e^0$ and passes it through a recurrent decoder followed by a dense decoder and an output layer that finally generates a sequence of predicted future velocities $\vv_{q}^{1:T_H}$ for the query robot over the prediction horizon.

\subsection{Model Training}
Using the generated demonstration data in Section \ref{subsec:dataset}, the designed model is trained end-to-end using back-propagation through time (BTTP) \cite{werbos1990backpropagation} with a fixed truncation depth $t_{\tn{trunc}}$. We learn the trajectory prediction model by minimizing the following loss function,
\begin{equation} \label{eq:mse}
	\begin{aligned}
		L(\vv_q^{1:T_H}, \boldsymbol\theta) = \frac{1}{T_H}\sum_{k=1}^{T_H}\norm{\vv_q^{k} - \vv_{q, \tn{true}}^{k}}^2 + \lambda\cdot l(\boldsymbol\theta),
		\end{aligned}
\end{equation}
where $\vv_{q,\tn{true}}^k$ is the ground truth velocity from the demonstration dataset, $l(\boldsymbol\theta)$ represents the regularization terms and $\lambda$ is the regularization factor. In our model, the $L2$ regularization method is adopted. 


\subsection{Decentralized Multi-Robot Motion Planning}
Having the trained trajectory prediction model, we can incorporate it with the MPC framework and solve the problem (\ref{eq:dmpc}) in a decentralized manner. As shown in Fig. \ref{fig:overview}, in a multi-robot navigation scenario, each robot first performs inference with the trained neutral network to predict the future trajectories of its neighboring robots and then plans a collision-free trajectory accordingly. Hence, decentralized multi-robot motion planning in dynamic environments is achieved. To be able to perform the inference, each robot needs to measure its own state as well as its neighbors', and keep a history memory of the information for a time horizon $T_O$. In addition, the robot also needs to measure the current states of moving obstacles in the environment.

\section{Results}\label{sec:result}
We now present results of simulation comparing the proposed approach with other methods and real-world experiments with quadrotors. 
A video accompanying this paper includes additional simulation and experimental results.

\subsection{Implementation Details}
To generate the dataset, we use an existing MATLAB multi-robot collision avoidance simulator\footnote{Code: \href{https://github.com/tud-amr/mrca-mav}{\color{blue}{https://github.com/tud-amr/mrca-mav}}} \cite{Zhu2019RAL} and simulate $N_{\tn{sim}} = 10^5$ time steps in a $10\times 10 \times 3$ m environment with \rebuttal{10 robots and 10 moving obstacles}. The robot we simulate is the Parrot Bebop 2 quadrotor with a radius set as 0.4 m. 
Ellipsoids representing the moving obstacles have semi-axes $(0.4, 0.4, 0.9)$ m. The sampling time and MPC planning horizon length are $\Delta t = 0.05$ s and $N = 20$, respectively. 
\rebuttal{
    We employ the same dynamics model and cost functions in the MPC problem (2) of our previous work \cite{Zhu2019RAL}.}
The Forces Pro solver \cite{domahidi2014forces} is used to solve the MPC problem. We set $T_O = 20$ and $T_H = 20$ the horizon length for robot past states observation and trajectory prediction. We further generate another test dataset by running the simulator in \rebuttal{six} different scenarios for $2\times 10^4$ time steps for each one of them. 
\rebuttal{
    All computations are performed in a commodity computer with an Intel i7 CPU@2.60GHz and an NVIDIA GTX 1060 GPU.}

The designed learning network is implemented in Python using TensorFlow 2. All layers in the network have 64 neurons except for the recurrent decoder that has 128 neurons and the output layer that has 3 neurons. While the activation function of the output layer is linear, all other layers in the network use a hyperbolic activation function. The regularization factor used during model training is $\lambda = 0.01$.

\begin{figure*}[t]
    \centering
    \setlength{\textfloatsep}{0.2\baselineskip plus 0.0\baselineskip minus 0.4\baselineskip}
    \captionsetup[subfigure]{position=b}
    \begin{subfigure}{0.195\textwidth}
            \includegraphics[width=1.0\textwidth]{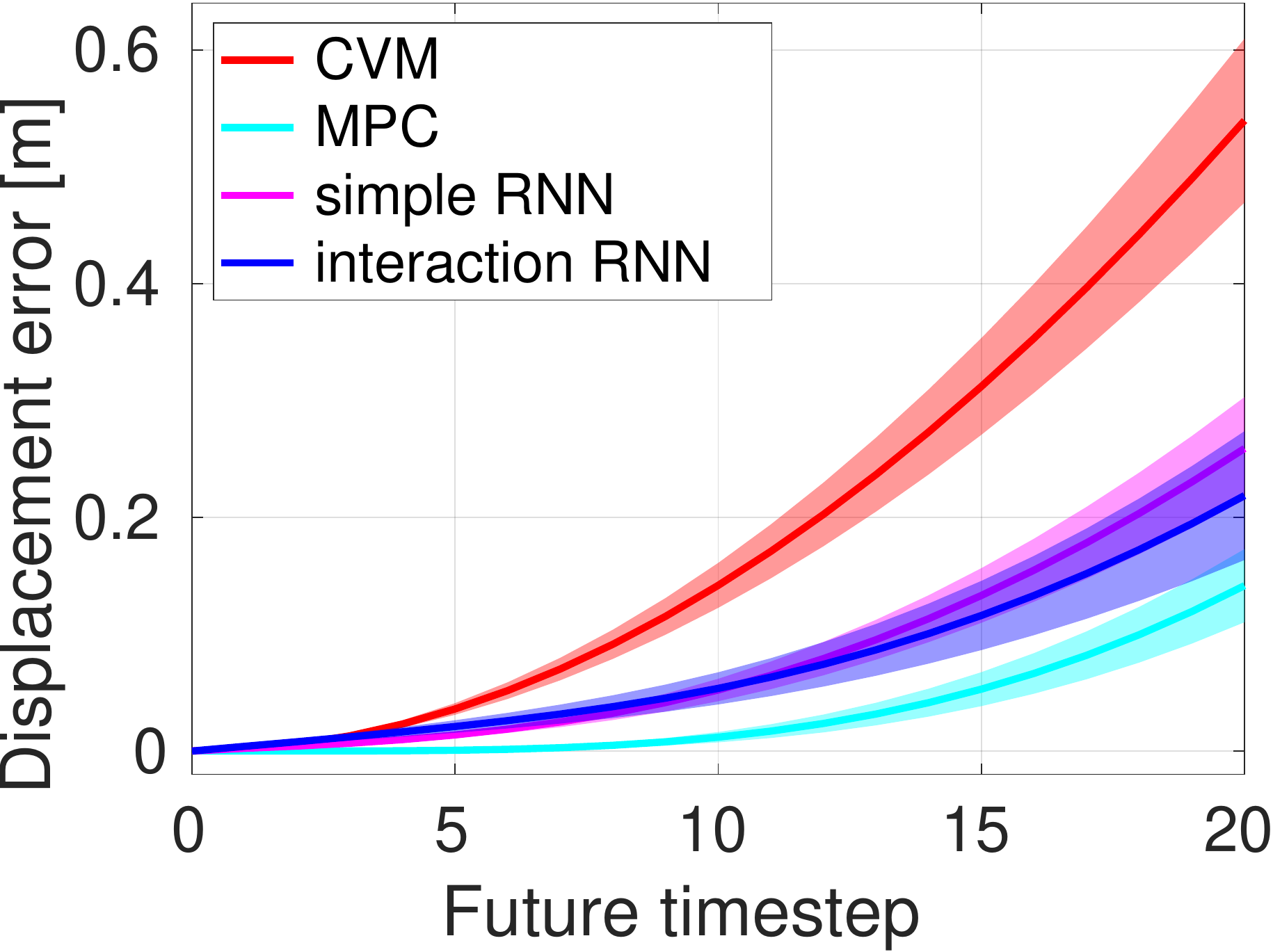}
    \caption{4 quads, no obstacles.}
	\end{subfigure}
    \captionsetup[subfigure]{position=b}
    \begin{subfigure}{0.195\textwidth}
            \includegraphics[width=1.0\textwidth]{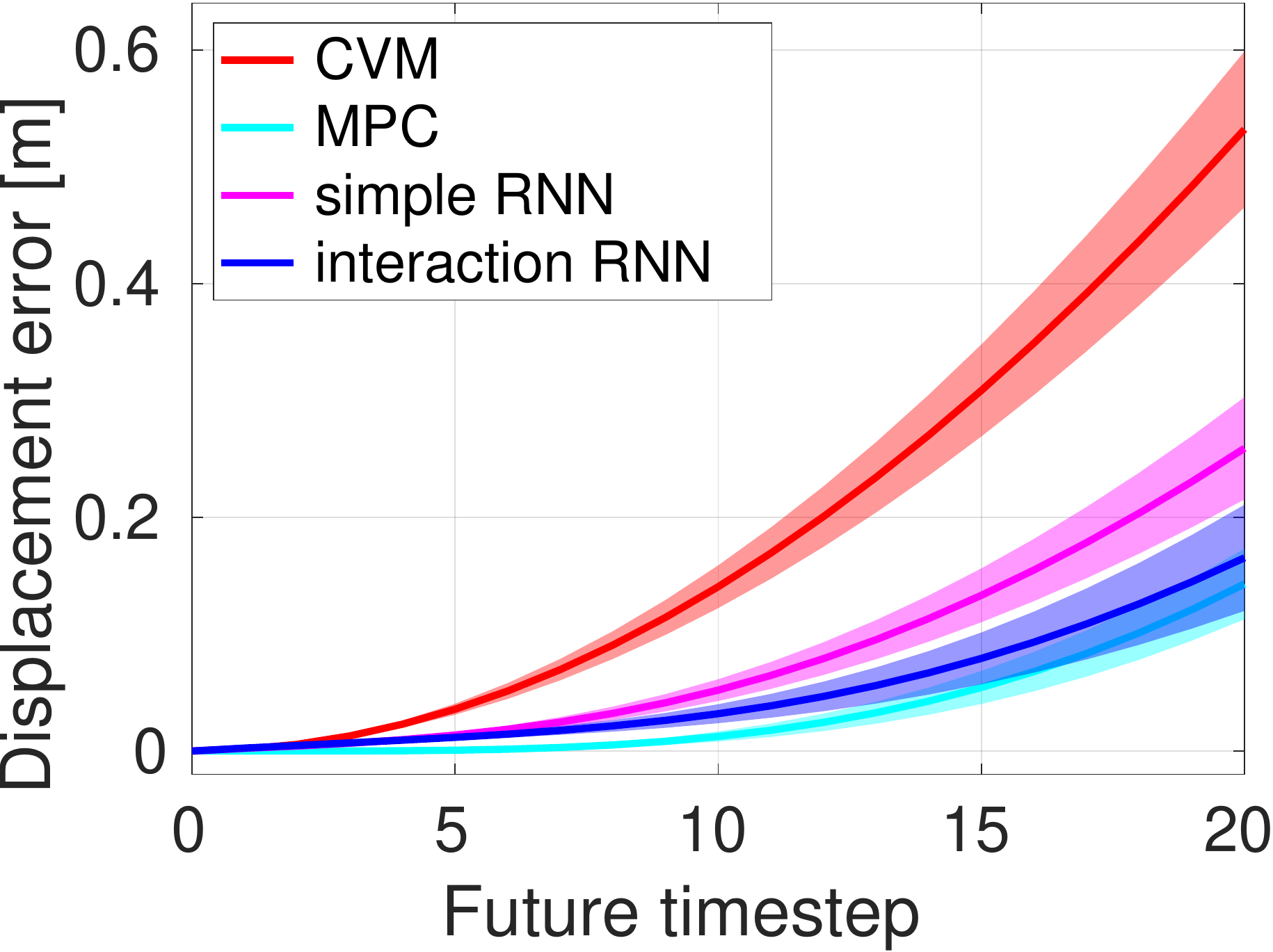}
    \caption{10 quads, no obstacles.}
	\end{subfigure}
    \captionsetup[subfigure]{position=b}
    \begin{subfigure}{0.195\textwidth}
            \includegraphics[width=1.0\textwidth]{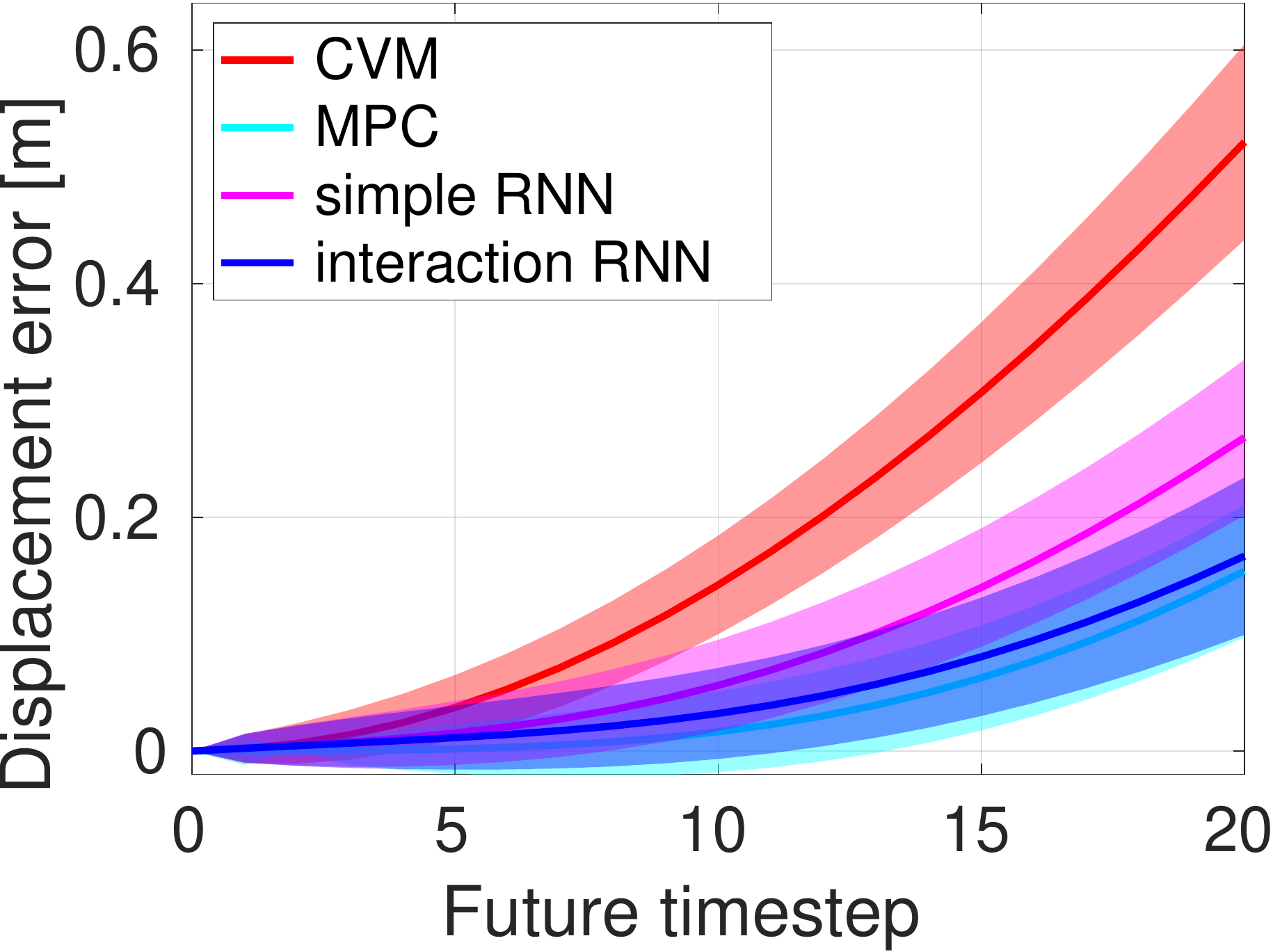}
    \caption{10 quads, 10 obstacles.}
    \end{subfigure}
    \captionsetup[subfigure]{position=b}
    \begin{subfigure}{0.195\textwidth}
            \includegraphics[width=1.0\textwidth]{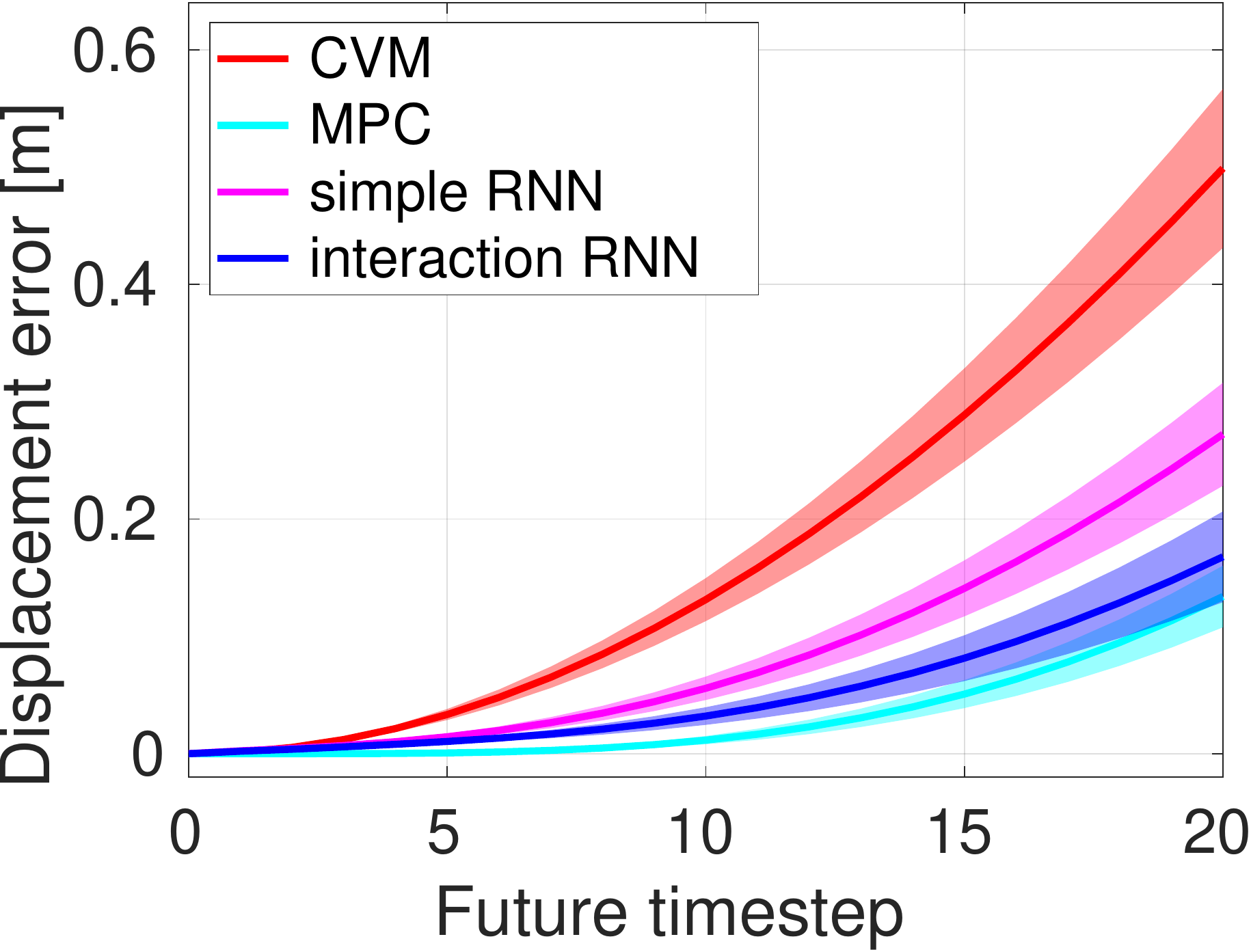}
    \caption{\rebuttal{20 quads, no obstacles.}}
	\end{subfigure}
    \captionsetup[subfigure]{position=b}
    \begin{subfigure}{0.195\textwidth}
            \includegraphics[width=1.0\textwidth]{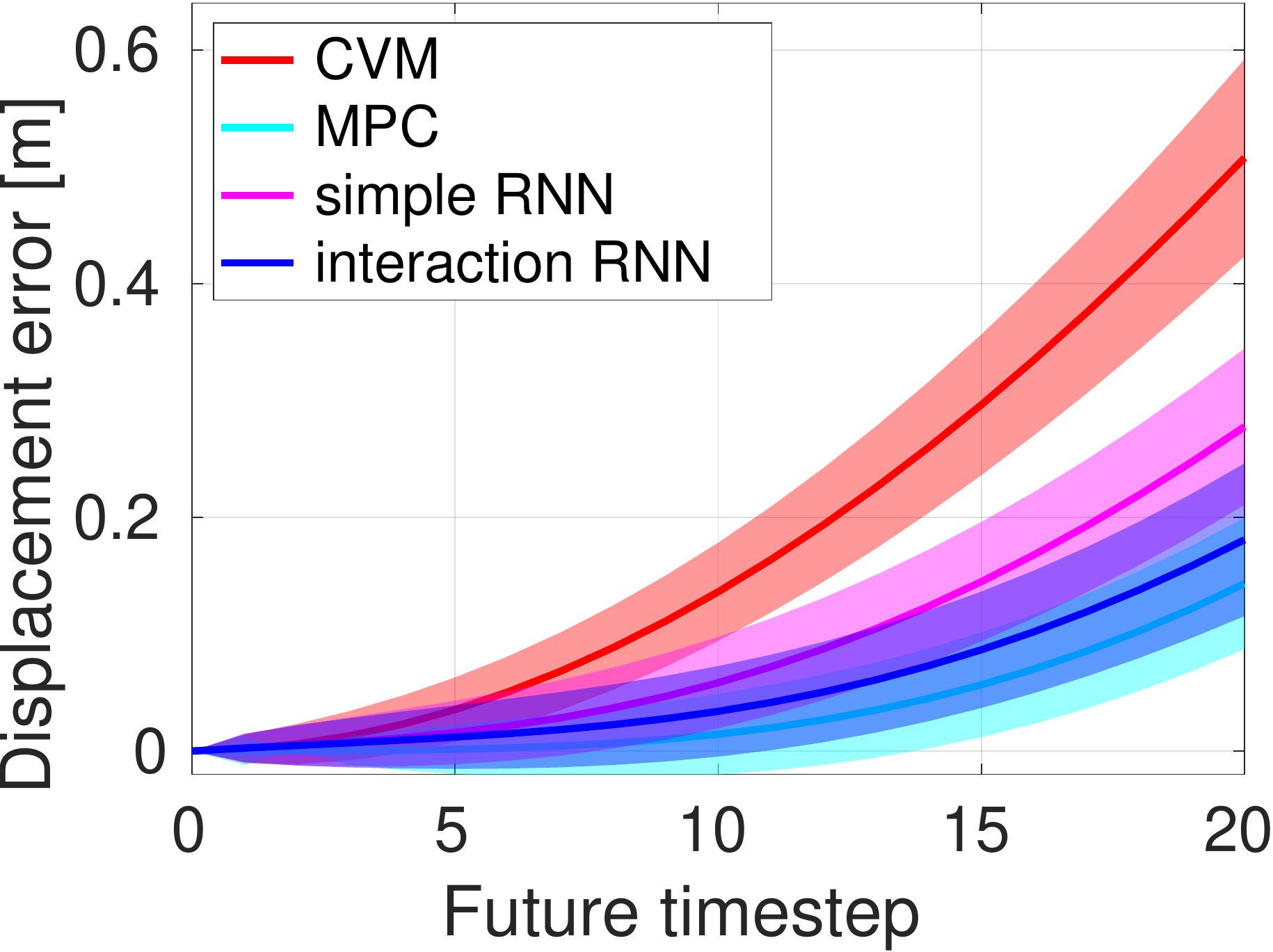}
    \caption{\rebuttal{20 quads, 10 obstacles.}}
    \end{subfigure}
    \caption{Performance results of our proposed interaction-aware RNN model for trajectory prediction compared to the baselines. The solid lines represent the average errors along the prediction horizon and the filled patches around them are 30\% of the standard deviation. The sampling period is 50 ms and the prediction horizon has 20 timesteps. }
    \label{fig:prediction_results}%
\end{figure*}

\subsection{Trajectory Prediction Evaluation}
We first evaluate our trajectory prediction model on a test dataset that has not been used for training nor validation. The dataset includes different test scenarios: an open environment with 4, 10, and \rebuttal{20} quadrotors, and with 10 moving obstacles. We compare our interaction-aware RNN-based model to three alternative methods: a) the constant velocity model (CVM) that is widely used in decentralized multi-robot motion planning; b) a simple RNN model that only considers the query robot's past states for trajectory prediction while ignoring its surrounding environment (this allows us to highlight the interaction awareness of our designed model); and c) an open-loop MPC planner assuming that the goal, robot model and constraints are known.

In Fig. \ref{fig:prediction_results} we present quantitative results of the prediction error with respect to ground truth in the test dataset. Recall that ground truths are the recorded robot traveled trajectories computed with the centralized sequential MPC (closed-loop). As expected, the prediction error of the open-loop MPC has the smallest prediction error among the methods since it was used for data generation and has perfect knowledge about the goal locations of all robots, which are not available for prediction in our proposed RNN-based model. Our proposed model can still achieve accurate trajectory predictions and significantly outperforms the CVM method across all scenarios. Moreover, compared to the simple RNN model, our interaction-aware approach achieves more accurate trajectory predictions, particularly in cluttered scenarios where interactions among robots are more frequent, as shown in Fig. \ref{fig:prediction_results}(b)-(e).
\rebuttal{
Furthermore, to evaluate the generalization capability of the learned network, we perform simulations in the scenarios (d) and (e) with 20 quadrotors which are beyond our training dataset. The results show that the proposed model still performs well on trajectory prediction in the two scenarios.}

\subsection{Decentralized Motion Planning}
We then evaluate performance of the proposed decentralized planner that incorporates the learned trajectory prediction model. 
\subsubsection{\rebuttal{Comparisons to other methods}}
We compare our method to the centralized sequential planning method \cite{Zhu2019RAL} with full communication among robots and the decentralized planning method \cite{Kamel2017} that uses the constant velocity model (CVM) for trajectory prediction to analyze whether more accurate trajectory forecasts of our RNN-based model lead to better planning performance. \rebuttal{Besides, another decentralized method, the buffered Voronoi cell (BVC) \cite{Zhou2017}, which guarantees collision avoidance is also implemented for comparison.}

Six quadrotors flying in four types of scenarios that represents different levels of difficulty \cite{Serra2020} are considered. Moreover, in order to avoid potential bias results, each scenario includes 50 instances where the robots have different starting and goal locations. The four scenarios are: \emph{1) symmetric swap}, in which the robots initially located at the vertices of a virtual horizontal regular hexagon are required to exchange their positions; \emph{2) asymmetric swap}, which differs from the previous scenario in that the hexagons are irregular, thus leading to more challenging collision-avoidance problems; \emph{3) pair-wise swap}, in which the robots are placed at random starting positions and assigned to three pairs within which the two robots need to swap their positions; and \emph{4) random moving}, in which each robot moves from a random starting position to a random goal in the environment.

\begin{figure}[t]
	\centering
    \captionsetup[subfigure]{position=b}
    \begin{subfigure}{0.151\textwidth}
            \includegraphics[width=1.0\textwidth]{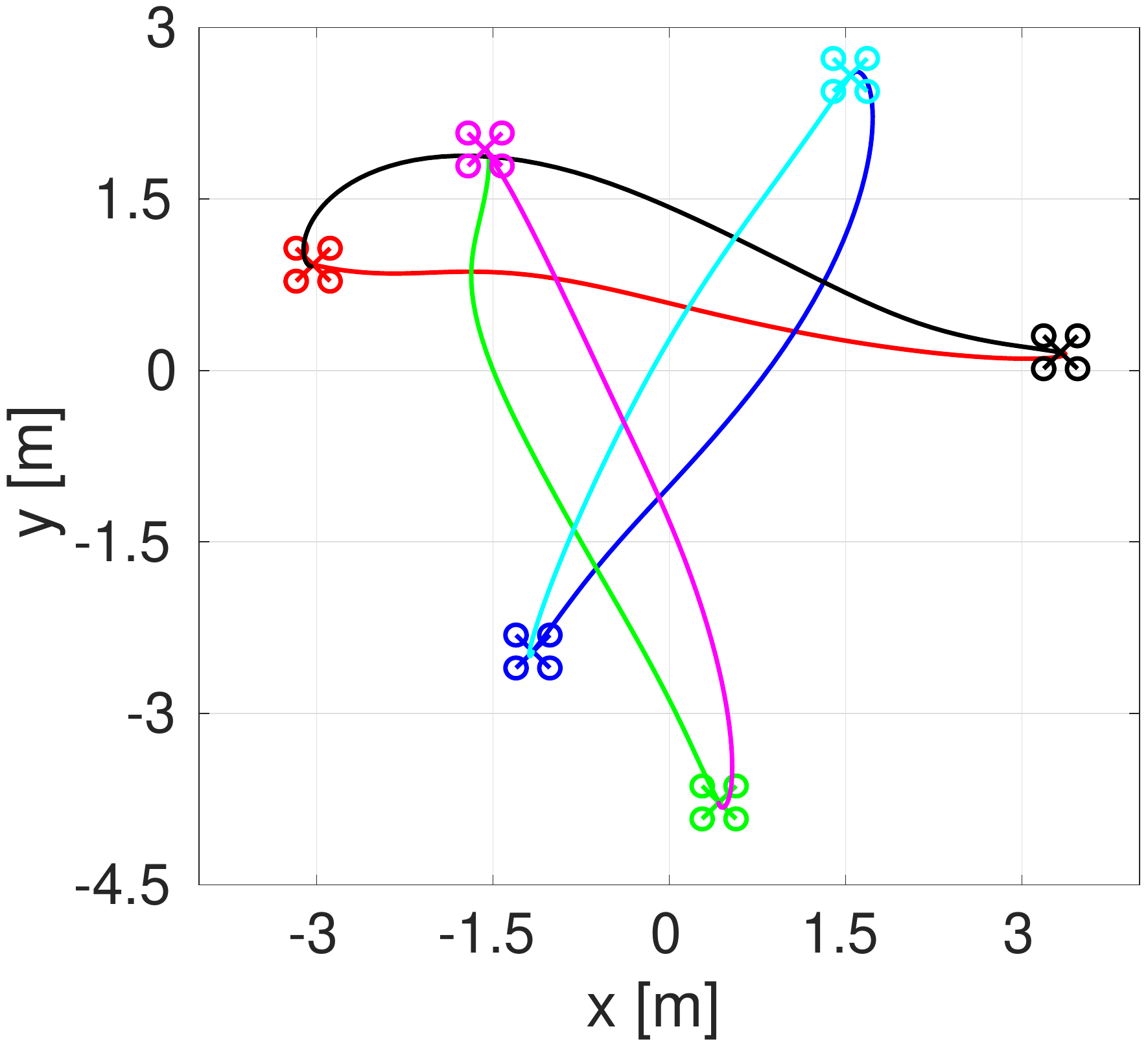}
	\end{subfigure}
    \captionsetup[subfigure]{position=b}
    \begin{subfigure}{0.151\textwidth}
            \includegraphics[width=1.0\textwidth]{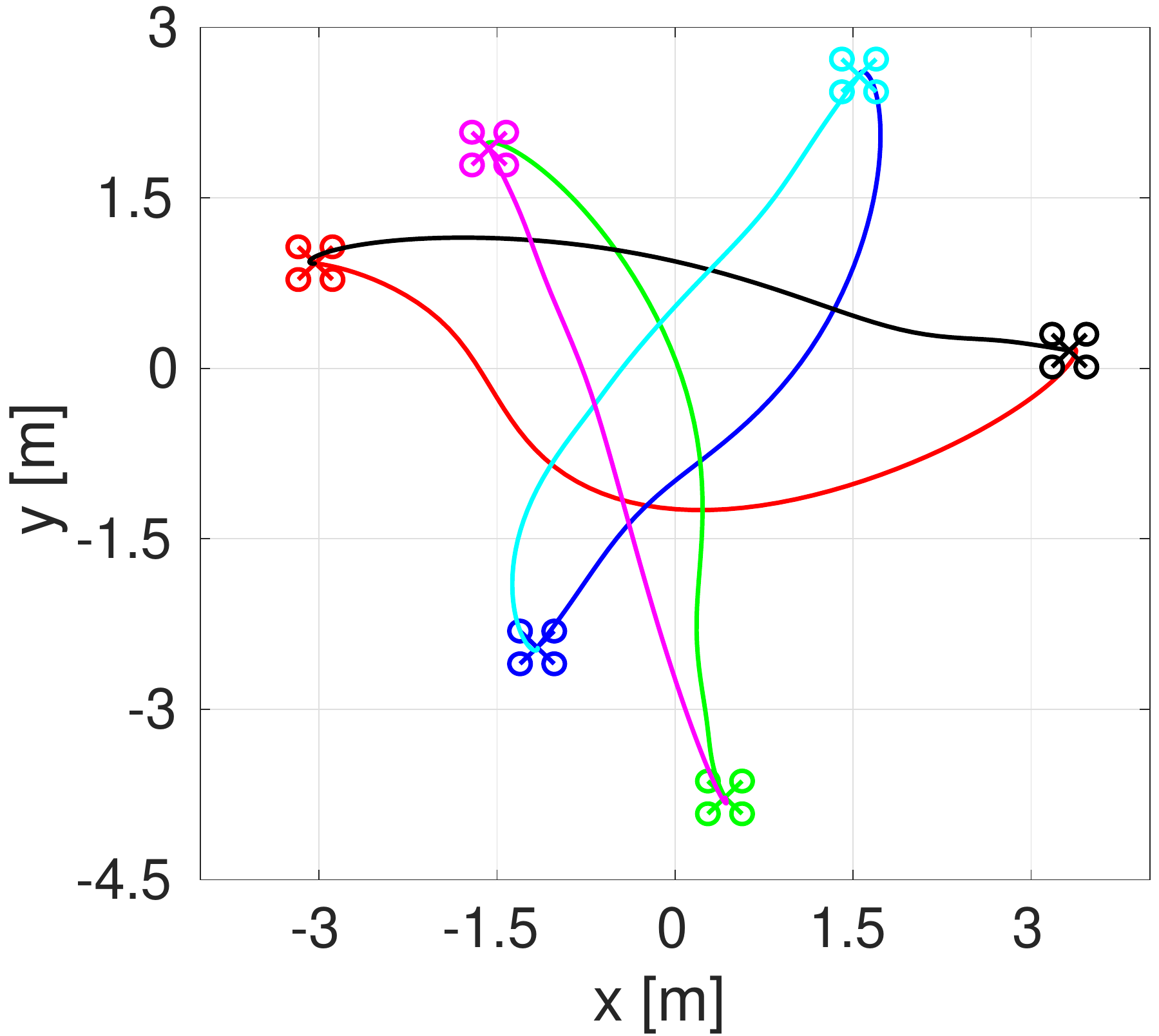}
	\end{subfigure}
	\captionsetup[subfigure]{position=b}
    \begin{subfigure}{0.151\textwidth}
            \includegraphics[width=1.0\textwidth]{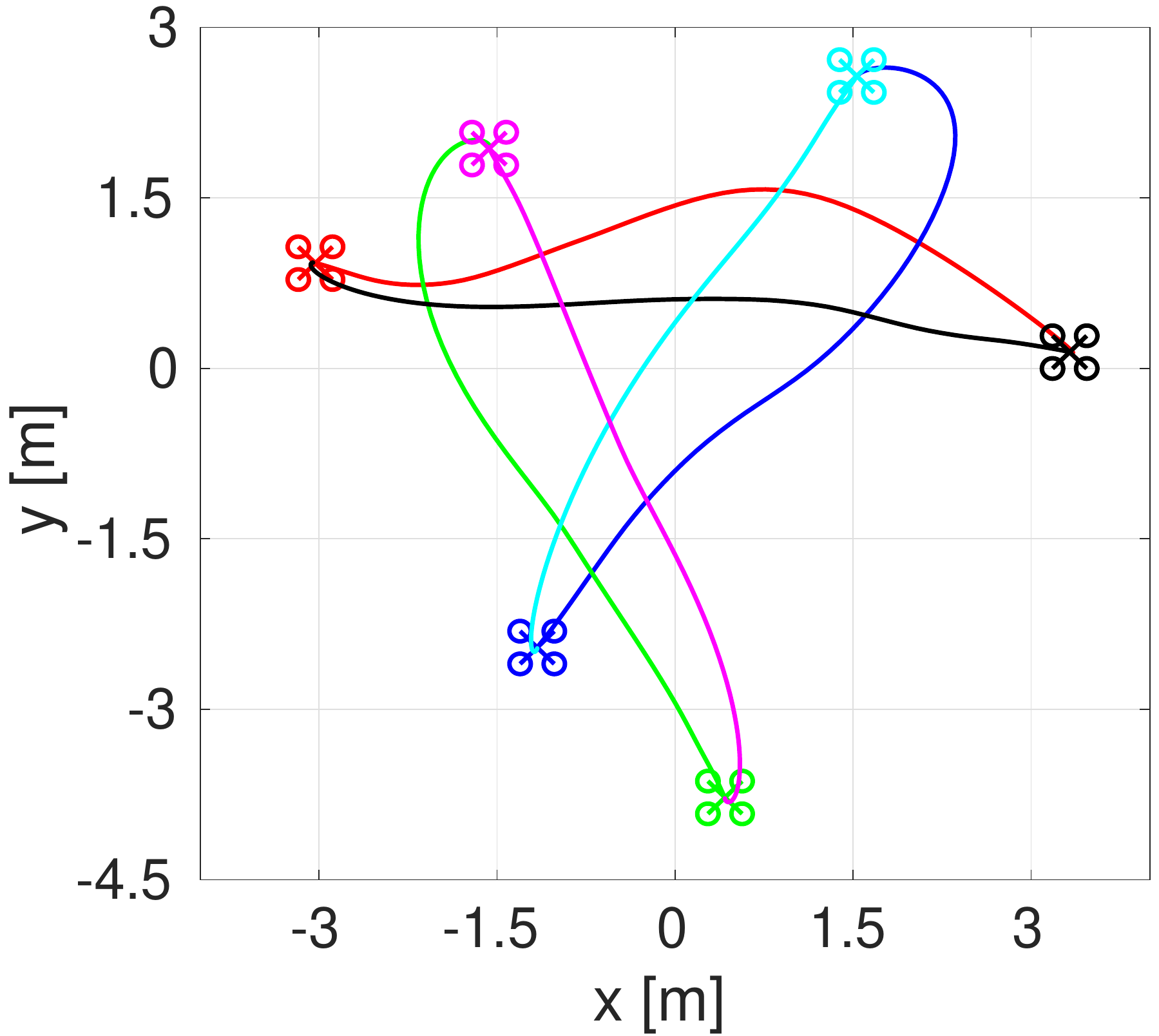}
	\end{subfigure}
	\\
	\captionsetup[subfigure]{position=b}
    \begin{subfigure}{0.15\textwidth}
            \includegraphics[width=1.0\textwidth]{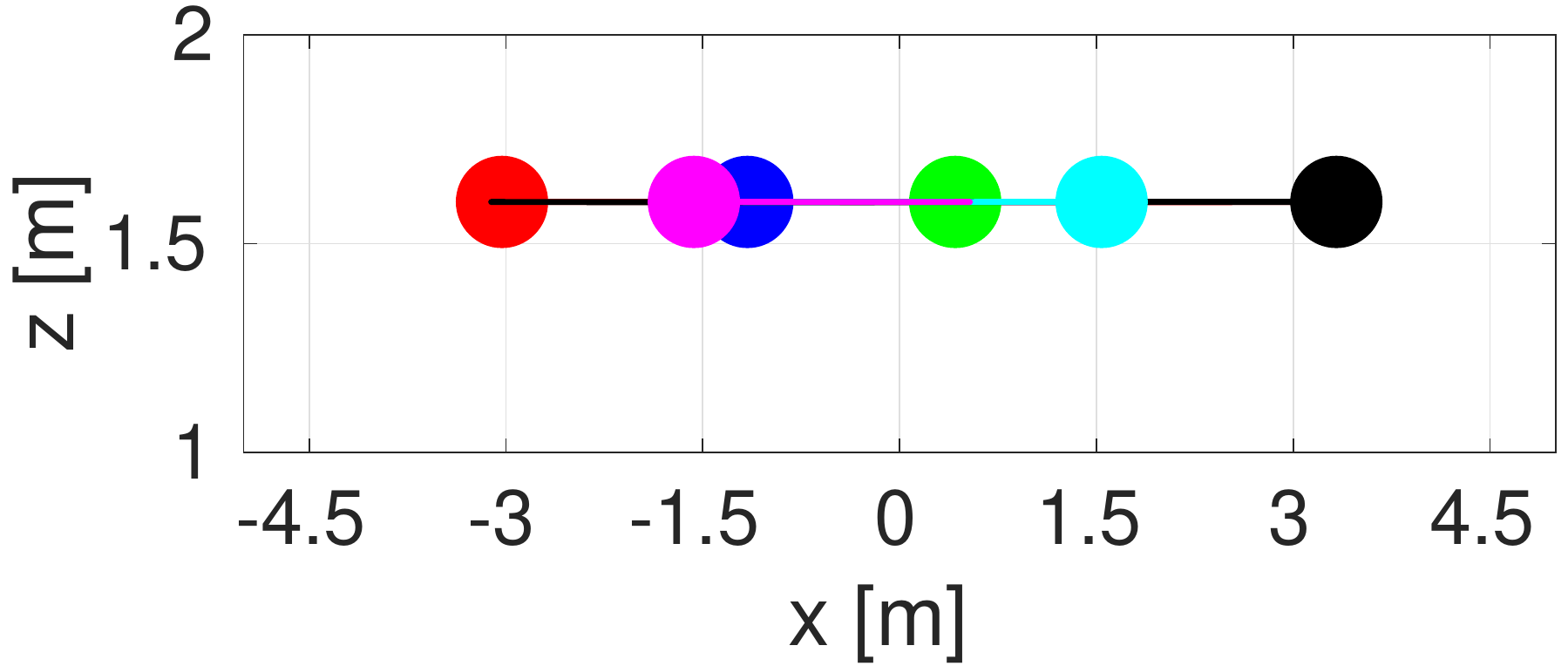}
	\caption{Cen. Comm.}
	\end{subfigure}
    \captionsetup[subfigure]{position=b}
    \begin{subfigure}{0.15\textwidth}
            \includegraphics[width=1.0\textwidth]{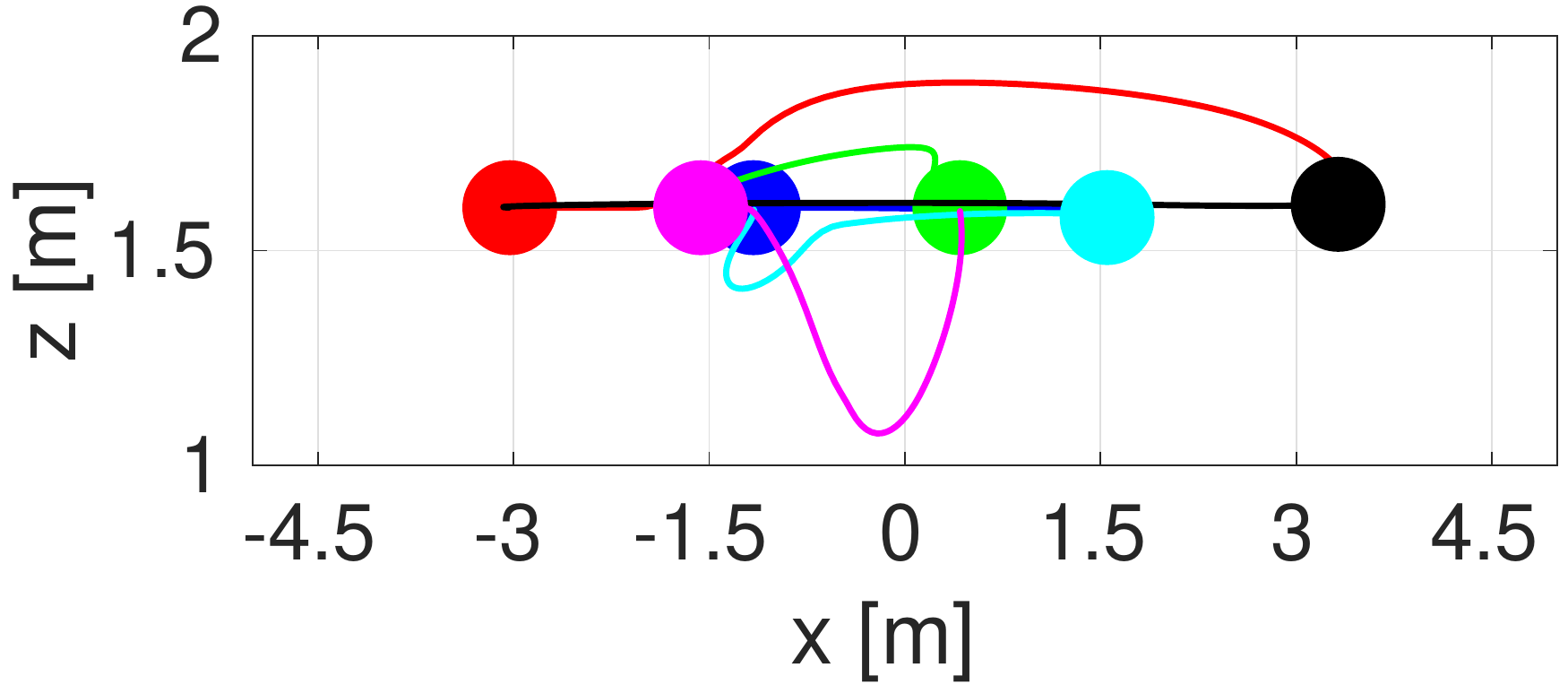}
	\caption{Decen. CVM}
	\end{subfigure}
	\captionsetup[subfigure]{position=b}
    \begin{subfigure}{0.15\textwidth}
            \includegraphics[width=1.0\textwidth]{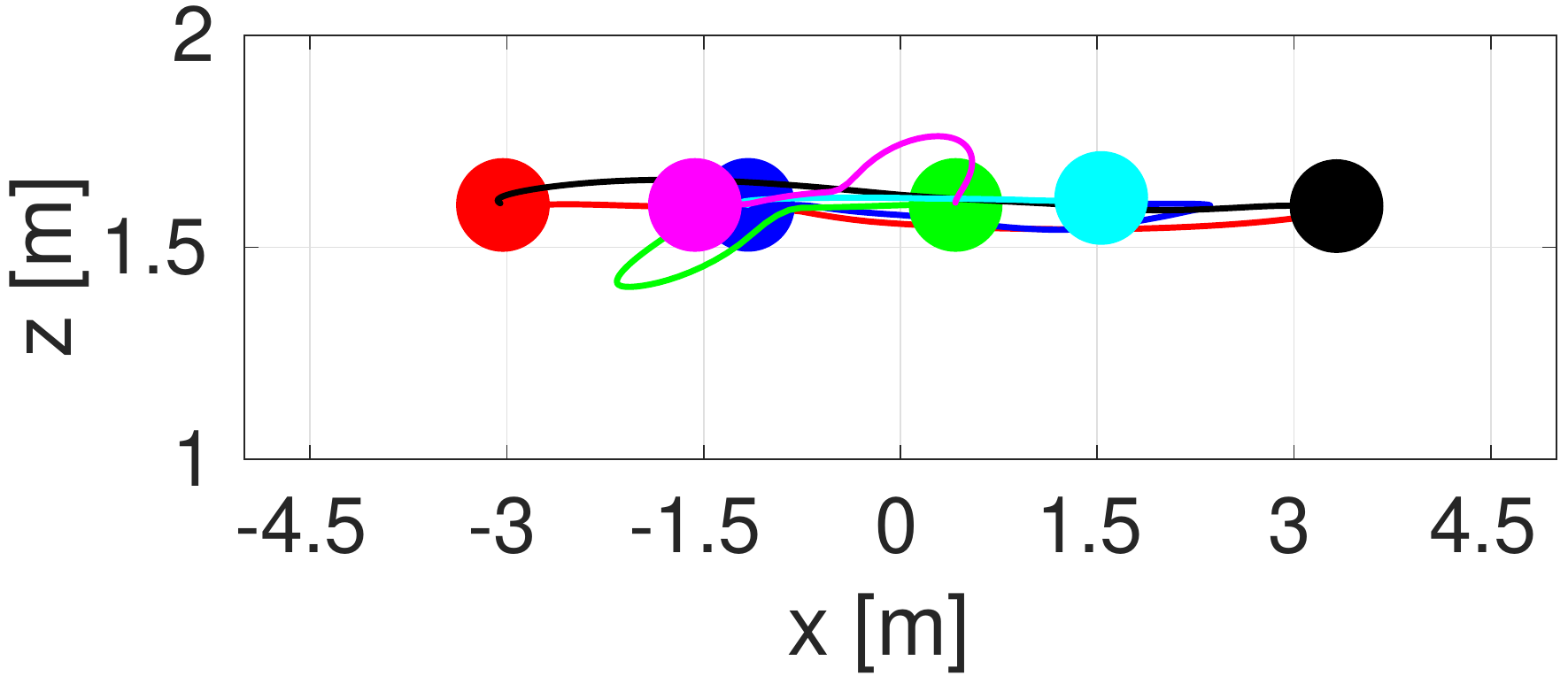}
	\caption{Decen. RNN}
	\end{subfigure}
    \caption{Simulation results of six quadrotors exchanging positions in the asymmetric swap scenario. Solid lines represent the trajectories. The upper and lower plots show the top view (X-Y) and side view(X-Z), respectively.}
    \label{fig:sim_planning}
\end{figure}

Qualitatively, Fig. \ref{fig:sim_planning} shows the sample trajectory trails of the six quadrotors for one instance from the asymmetric swap scenario. \rebuttal{It can be seen that our RNN-based decentralized planner achieves results that are closer to the centralized sequential planner than the CVM-based planner.}
To quantitatively evaluate the performance of different motion planners, we consider a wide range of metrics: the number of instances that lead to collisions within the entire 50 runs, the average trajectory length and trajectory duration of the team of robots, and the overall robot average speed during the whole simulation. The last three metrics are only computed for those successful runs. Table \ref{tab:planning_comparison} summaries the simulation results. It can be seen that our RNN-based planner significantly outperforms the planner using the CVM for trajectory prediction in terms of safety, in particular in the challenging asymmetric swap scenario. In addition, \rebuttal{our planner also achieves consistently smaller trajectory lengths and durations compared to the CVM-based planner} in all scenarios. 
\rebuttal{Compared to the BVC method, our proposed approach achieves significantly shorter trajectory durations, particularly in the (a)symmetric swapping scenarios, which shows superiority of the MPC framework over the reactive BVC method.}
Finally, compared to the centralized sequential planner with full communication, our planner can achieve a comparable level of performance in terms of safety and trajectory efficiency while being decentralized and communication-free. However, three instances out of 50 in the challenging asymmetric swap scenario is still observed with collisions using the RNN-based method, indicating that in few rare cases, highly-accurate trajectory predictions of other robots, for example obtained via communication, are necessary to ensure safety. 
\rebuttal{
    In the simulation, on average the computation time of the proposed decentralized MPC planner with the learned predictor is 36.3 ms, which is smaller than that of the centralized sequential planner which plans trajectories for all six robots (43.9 ms). Besides, our decentralized approach is communication-free.}


\begin{table*}[t]
	\centering
	\caption{Performance comparison of different multi-robot motion planners (centralized with communication \cite{Zhu2019RAL}, \rebuttal{decentralized buffered Voronoi cell (BVC) method \cite{Zhou2017}}, decentralized with constant velocity model (CVM) \cite{Kamel2017} and decentralized with our RNN-based model) across the four different types of scenarios (symmetric swap, asymmetric swap, pair-wise swap and random moving). Each scenario includes 50 running instances. }
    \begin{tabular}{ll||c|ccc|ccc|c}
        \hline
		\multicolumn{1}{c}{\multirow{2}{*}{\textbf{Scenario}}}                     & \multicolumn{1}{c||}{\multirow{2}{*}{\textbf{Motion Planner}}} & \multirow{2}{*}{\textbf{\begin{tabular}[c]{@{}c@{}}Num. of coll. \\ instances\end{tabular}}} & \multicolumn{3}{c|}{\textbf{Trajectory length (m)}} & \multicolumn{3}{c|}{\textbf{Trajectory duration (s)}} & \multirow{2}{*}{\textbf{\begin{tabular}[c]{@{}c@{}}Average speed\\ (m/s)\end{tabular}}} \\
		\multicolumn{1}{c}{}                                                       & \multicolumn{1}{c||}{}                                         &                           & Min.                          & Average                   & Max.                  & Min.                  & Average                       & Max.                      &                                   \\ \hline \hline
        \multirow{3}{*}{\begin{tabular}[c]{@{}l@{}}Symmetric\\ swap\end{tabular}}  & Cen. Comm.                                                     & {0}                       & 5.46                          & 7.22$\pm$0.85     	    & 8.96                  & 5.15                  & 5.62$\pm$0.24                 & 6.05                      & 1.27$\pm$0.05                     \\
                                                                                   & \rebuttal{Decen. (BVC)}                                        & \tb{\rebuttal{0}}         & \rebuttal{5.75}               & \rebuttal{7.63$\pm$0.87}  & \rebuttal{9.63}       & \rebuttal{10.00}      & \rebuttal{12.21$\pm$1.25}     & \rebuttal{14.80}          & \rebuttal{0.63$\pm$0.03}          \\
																				   & Decen. (CVM)                                                   & 4                         & 5.43                          & 7.45$\pm$0.88             & 9.55                  & 4.95                  & 6.02$\pm$0.55                 & 7.30                      & 1.23$\pm$0.05                     \\
																				   & {Decen. (RNN)}                                                 &  \tb{{0}}                 &  {5.41}                       &  \tb{7.35$\pm$0.91}       &  {10.60}              &  {4.75}               &  \tb{5.59$\pm$0.40}           &  {9.70}                   &  \tb{1.30$\pm$0.05}               \\ \hline
        \multirow{3}{*}{\begin{tabular}[c]{@{}l@{}}Asymmetric\\ swap\end{tabular}} & Cen. Comm.                                                     & {0}                       & 5.08                          & 6.77$\pm$0.80             & 9.02                  & 4.65                  & 5.17$\pm$0.30                 & 5.85                      & 1.30$\pm$0.05                     \\
                                                                                   & \rebuttal{Decen. (BVC)}                                        & \tb{\rebuttal{0}}         & \rebuttal{5.31}               & \rebuttal{7.25$\pm$0.87}  & \rebuttal{9.29}       & \rebuttal{9.70}       & \rebuttal{11.34$\pm$1.00}     & \rebuttal{13.90}          & \rebuttal{0.65$\pm$0.03}          \\
																				   & Decen. (CVM)                                                   & 15                        & 5.32                          & 7.76$\pm$1.89             & 18.06                 & 5.05                  & 5.96$\pm$0.51                 & 7.30                      & 1.28$\pm$0.04                     \\
																				   & {Decen. (RNN)}                                                 &  {3}                      &  {5.16}                       &  \tb{7.14$\pm$1.10}       &  {12.60}              &  {4.80}               &  \tb{5.48$\pm$0.42}           &  {6.85}                   &  \tb{1.29$\pm$0.05}               \\ \hline
		\multirow{3}{*}{\begin{tabular}[c]{@{}l@{}}Pair-wise\\ swap\end{tabular}}  & Cen. Comm.                                                     & {0}                       & 1.64                          & 5.10$\pm$1.87             & 9.92                  & 3.50                  & 4.76$\pm$0.44                 & 5.85                      & 1.06$\pm$0.12                     \\
                                                                                   & \rebuttal{Decen. (BVC)}                                        & \tb{\rebuttal{0}}         & \rebuttal{1.83}               & \rebuttal{5.25$\pm$1.93}  & \rebuttal{10.13}      & \rebuttal{5.20}       & \rebuttal{7.58$\pm$1.66}      & \rebuttal{13.20}          & \rebuttal{0.67$\pm$0.07}          \\
                                                                                   & Decen. (CVM)                                                   & 3                         & 1.74                          & 5.54$\pm$2.53             & 17.42                 & 3.60                  & 5.00$\pm$0.65                 & 5.75                      & \tb{1.06$\pm$0.10}                \\
																				   & {Decen. (RNN)}                                                 &  \tb{0}                   &  {1.70}                       &  \tb{4.94$\pm$2.02}       &  {9.94}               &  {3.40}               &  \tb{4.83$\pm$0.45}           &  {5.95}                   &  {1.01$\pm$0.14}                  \\ \hline
		\multirow{3}{*}{\begin{tabular}[c]{@{}l@{}}Random\\ moving\end{tabular}}   & Cen. Comm.                                                     & {0}                       & 0.39                          & 4.66$\pm$1.94             & 8.63                  & 3.50                  & 7.72$\pm$0.40                 & 5.50                      & 0.98$\pm$0.13                     \\
                                                                                   & \rebuttal{Decen. (BVC)}                                        & \tb{\rebuttal{0}}         & \rebuttal{0.39}               & \rebuttal{4.81$\pm$2.00}  & \rebuttal{8.84}       & \rebuttal{4.70}       & \rebuttal{7.13$\pm$1.30}      & \rebuttal{10.10}          & \rebuttal{0.69$\pm$0.09}          \\
                                                                                   & Decen. (CVM)                                                   & \tb{0}                    & 0.39                          & 4.82$\pm$2.06             & 9.53                  & 3.60                  & 4.89$\pm$0.57                 & 6.35                      & \tb{0.98$\pm$0.12}                \\
                                                                                   & {Decen. (RNN)}                                                 &  \tb{0}                   &  {0.39}                       &  \tb{4.36$\pm$2.11}       &  {9.19}               &  {3.85}               &  \tb{4.76$\pm$0.43}           &  {5.95}                   &  {0.91$\pm$0.11}                                                                                                                                  \\                                  
        \hline                        
	\end{tabular}
	\label{tab:planning_comparison}
\end{table*}

\rebuttal{


\subsubsection{Effect of non-MPC robots on performance}
Our proposed decentralized approach assumes that all robots interact and adopt the same motion planning strategy, namely MPC-based trajectory optimization with the learned motion prediction model. We now evaluate the performance of our approach in a mixture scenario where some robots employ the BVC method \cite{Zhou2017} for collision avoidance.
We simulate 50 instances with six quadrotors in the symmetric swap scenario of Section V-C-1. Table \ref{table:nonMPC_robot} presents the simulation results. When there is only one BVC robot, no collisions are observed. However, when more BVC robots are in the team, particularly when half of the robots (3) are BVC-based, collisions will happen due to incorrect motion predictions of them by other MPC robots. This indicates that the assumption that the robots interact with the same planning strategy is necessary to ensure safety. 

\begin{table}[t]
    \centering
    \caption{\rebuttal{Simulation results of six quadrotors in the symmetric swap scenario where a varying number of BVC-based robots are in the team. 50 running instances are simulated.}}
    \begin{tabular}{lcccccc}
    \hline
    Num. of BVC robots      & 0         & 1         & 2         & 3         & 4                  \\ \hline
    Num. of coll. instan.   & 0         & 0         & 2         & 4         & 1                  \\
    Ave. traj. time (s)     & 5.59      & 6.82      & 8.517     & 9.63      & 10.27              \\ \hline
    \end{tabular}
    \label{table:nonMPC_robot}
\end{table}

}

\subsection{Experimental Validation}
\subsubsection{Setup}
We validate our proposed approach with a team of Parrot Bebop 2 quadrotors flying in a shared space with walking human obstacles. The pose of each quadrotor and obstacle (human) is obtained using an external motion capture system (OptiTrack) and their velocities obtained via a standard Kalman filter running at a high rate. Control commands are sent to the quadrotor via ROS. 
\rebuttal{
    During the experiment, the humans walked at a speed with mean 0.8 m/s and the maximum 1.2 m/s. They could change their speeds and make small turns in the workspace. 
}

\begin{figure}[t]
    \centering
    ~
    \captionsetup[subfigure]{position=b}
    \begin{subfigure}{0.19\textwidth}
            \includegraphics[width=1.0\textwidth]{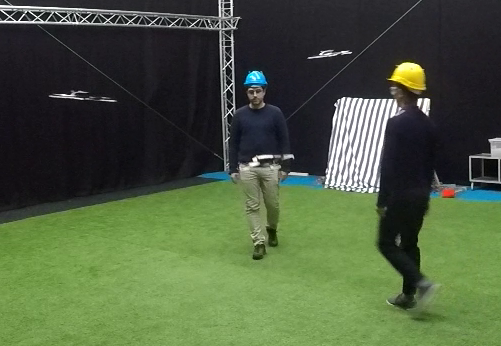}
	\caption{}
	\label{subfig:exp_gopro}
	\end{subfigure}
    \captionsetup[subfigure]{position=b}
    \begin{subfigure}{0.23\textwidth}
            \includegraphics[width=1.0\textwidth]{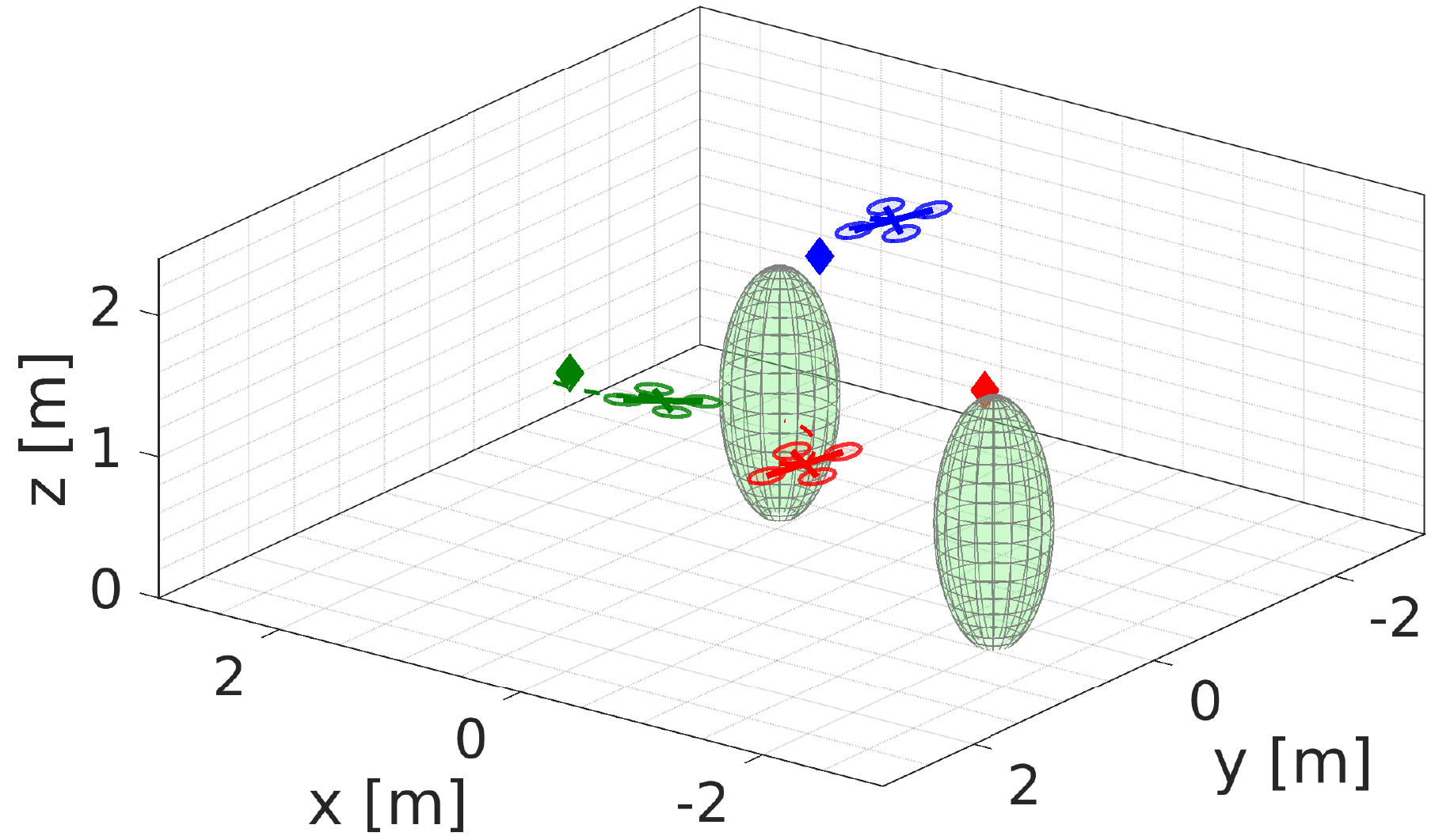}
	\caption{}
	\label{subfig:exp_matlab}
    \end{subfigure}
    \begin{subfigure}{0.21\textwidth}
            \includegraphics[width=1.0\textwidth]{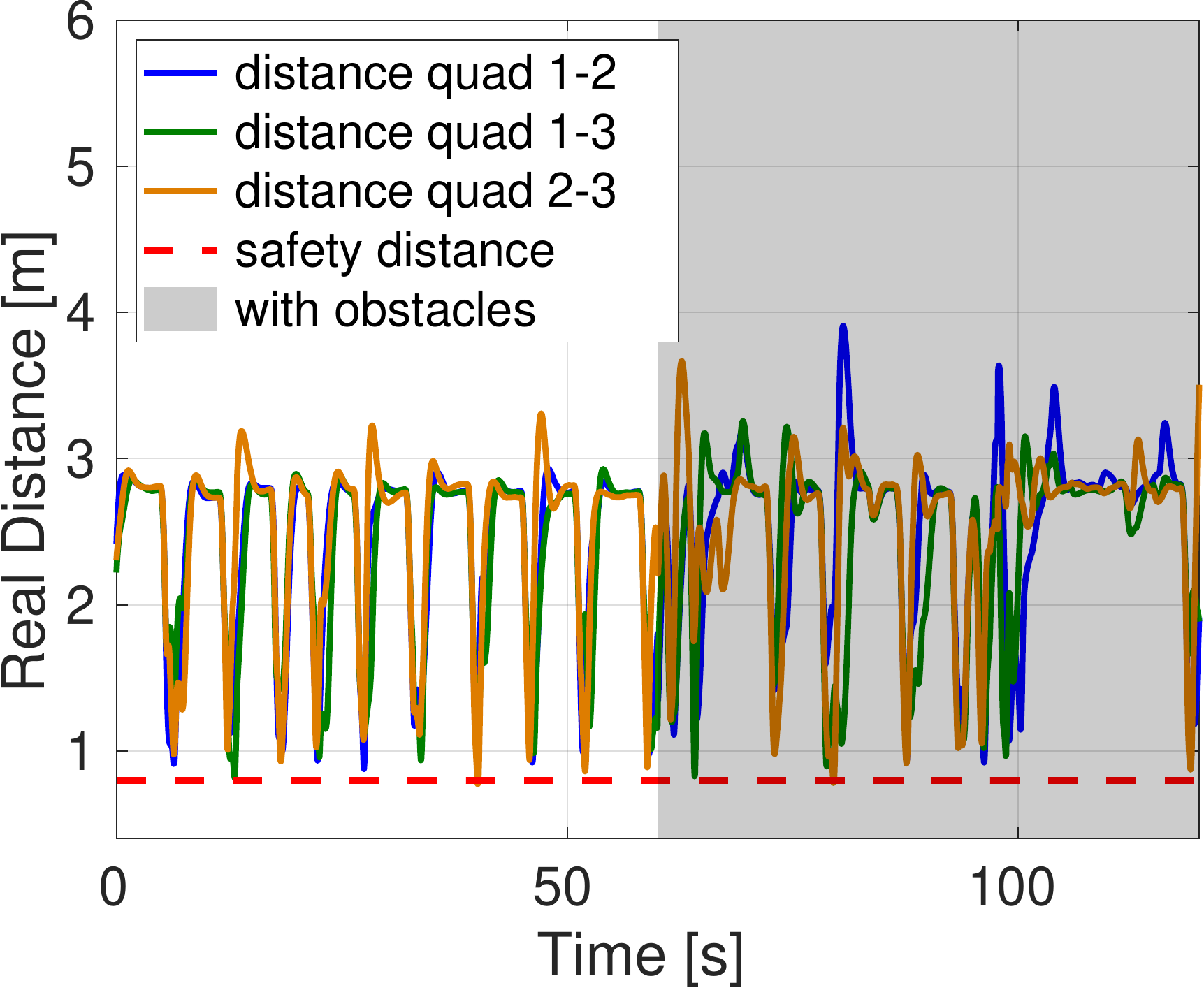}
	\caption{}
	\label{subfig:quad_dis}
	\end{subfigure}
    \captionsetup[subfigure]{position=b}
    \begin{subfigure}{0.23\textwidth}
            \includegraphics[width=1.0\textwidth]{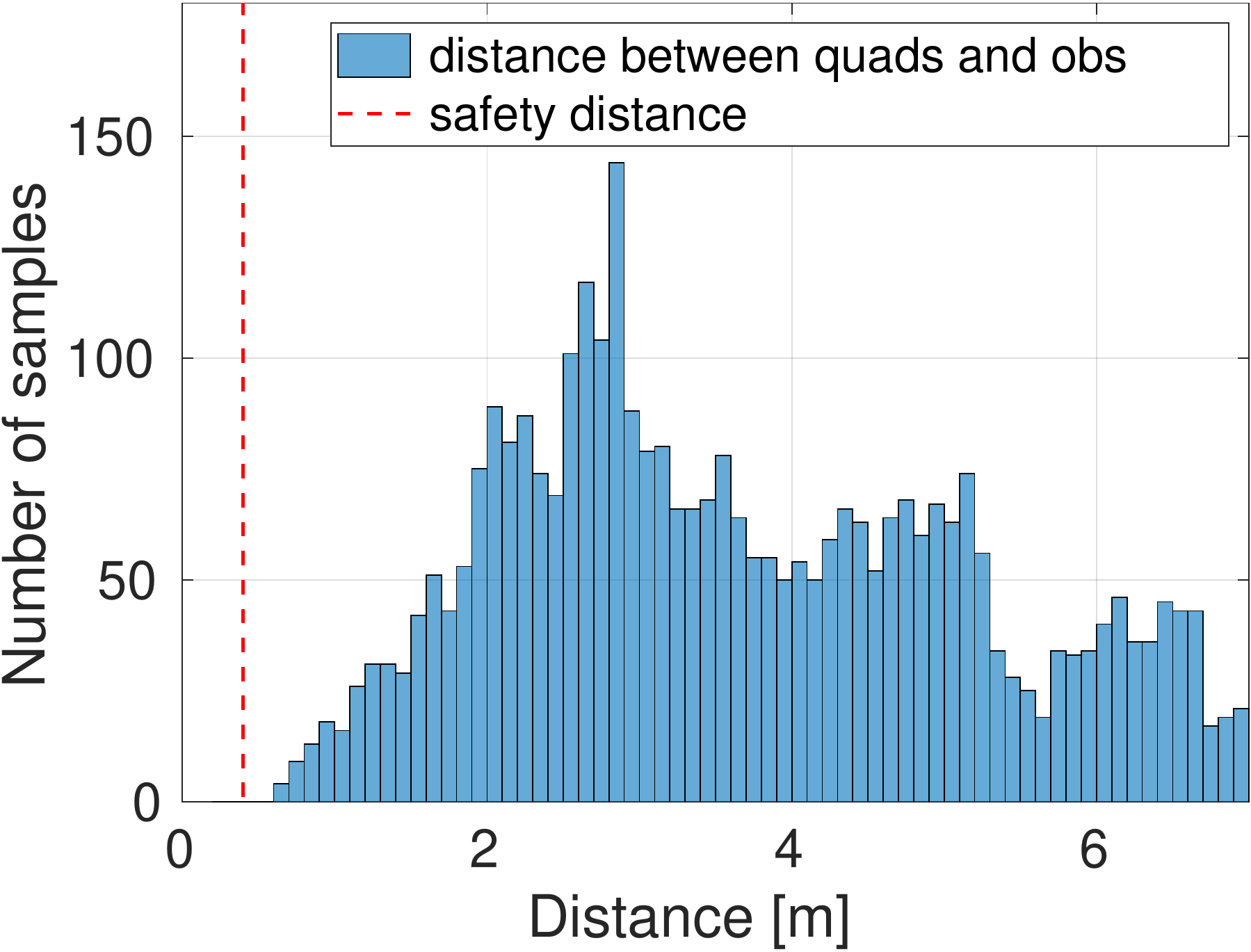}
	\caption{}
	\label{subfig:obs_dis}
    \end{subfigure}
    \caption{Experimental results with three quadrotors flying in a shared space with two walking humans. (a) A snapshot of the experiment. (b) Schematic of quadrotors, humans, and planned trajectories. (c) Distance between the quadrotors over time. The shaded grey area indicates the two walking humans join the space. (d) Histogram of the quadrotor-obstacle distance during the experiments.}
    \label{fig:exp_results}%
\end{figure}

\subsubsection{Results}
Experiments in two representative scenarios are conducted: with and without obstacles. In the first scenario, three quadrotors, initially distributed in a virtual horizontal circle, are required to swap their positions multiple times. Then in the second scenario, two moving obstacles (walking humans) join the space while the three quadrotors keeps changing their positions while avoiding the humans at the same time. Fig. \ref{subfig:exp_gopro} presents a snapshot from the experiment. 
Fig. \ref{subfig:quad_dis} shows distance between each pair of the three quadrotors over time during the experiment. It can be seen that they maintained a safe distance of 0.8 m over the entire run even after the two walking humans joined the space which makes it more confined. In Fig. \ref{subfig:obs_dis} we cumulate the distance between each quadrotor and human obstacle that is computed as the closest distance from the quadrotor center to the obstacle ellipsoid's surface. The results show that a minimum safe separation of 0.4 m to the obstacles is achieved. In sum, the demonstration shows that our proposed approach works well for multi-robot motion planning in dynamic environments which is decentralized and communication-free.

\section{Conclusion}\label{sec:conclsuion}
In this paper we presented a decentralized multi-robot MPC-based motion planning approach that accounts for the robot's interactions with obstacles and other robots through the use of a RNN-based trajectory prediction model. We showed that our proposed interaction-aware RNN model generalizes well with different number of robots and obstacles, and is able to provide more accurate trajectory predictions than the constant velocity model in a variety of scenarios. In simulation with six quadrotors, we showed that our decentralized planner outperforms the planner using a constant velocity model for trajectory prediction and can achieve a comparable level of performance to the centralized sequential planner while being communication-free. We also validated our approach in real-world experiments with three quadrotors flying in a shared space with walking humans. Future work shall take into account sensing uncertainties and consider more complex unstructured environments with static obstacles.

\bibliographystyle{IEEEtran}
\balance
\bibliography{ref_abb_2}

\begin{thebibliography}{10}
\providecommand{\url}[1]{#1}
\csname url@rmstyle\endcsname
\providecommand{\newblock}{\relax}
\providecommand{\bibinfo}[2]{#2}
\providecommand\BIBentrySTDinterwordspacing{\spaceskip=0pt\relax}
\providecommand\BIBentryALTinterwordstretchfactor{4}
\providecommand\BIBentryALTinterwordspacing{\spaceskip=\fontdimen2\font plus
\BIBentryALTinterwordstretchfactor\fontdimen3\font minus
  \fontdimen4\font\relax}
\providecommand\BIBforeignlanguage[2]{{%
\expandafter\ifx\csname l@#1\endcsname\relax
\typeout{** WARNING: IEEEtran.bst: No hyphenation pattern has been}%
\typeout{** loaded for the language `#1'. Using the pattern for}%
\typeout{** the default language instead.}%
\else
\language=\csname l@#1\endcsname
\fi
#2}}

\bibitem{breitenmoser2016combining}
A.~Breitenmoser and A.~Martinoli, ``{On combining multi-robot coverage and
  reciprocal collision avoidance},'' in \emph{Springer Tracts Adv. Robot.},
  2016, vol. 112, pp. 49--64.

\bibitem{baxter2007multi}
J.~L. Baxter, E.~K. Burke, J.~M. Garibaldi, and M.~Norman, ``{Multi-robot
  search and rescue: a potential field based approach},'' in \emph{Stud.
  Comput. Intell.}, 2007, vol.~76, pp. 9--16.

\bibitem{Zhu2019ICRA}
H.~Zhu, J.~Juhl, L.~Ferranti, and J.~Alonso-Mora, ``{Distributed multi-robot
  formation splitting and merging in dynamic environments},'' in \emph{Proc.
  IEEE Int. Conf. Robot. Autom. (ICRA)}, 2019, pp. 9080--9086.

\bibitem{Nageli2017}
T.~N{\"{a}}geli, L.~Meier, A.~Domahidi, J.~Alonso-Mora, and O.~Hilliges,
  ``{Real-time planning for automated multi-view drone cinematography},''
  \emph{ACM Trans. Graph.}, vol.~36, no.~4, pp. 1--10, 2017.

\bibitem{Zhu2019RAL}
H.~Zhu and J.~Alonso-Mora, ``{Chance-constrained collision avoidance for mavs
  in dynamic environments},'' \emph{IEEE Robot. Autom. Lett.}, vol.~4, no.~2,
  pp. 776--783, 2019.

\bibitem{Kamel2017}
M.~Kamel, J.~Alonso-Mora, R.~Siegwart, and J.~Nieto, ``{Robust collision
  avoidance for multiple micro aerial vehicles using nonlinear model predictive
  control},'' in \emph{Proc. IEEE/RSJ Int. Conf. Intell. Robot. Syst. (IROS)},
  2017, pp. 236--243.

\bibitem{vandenBerg2011}
J.~van~den Berg, S.~J. Guy, M.~Lin, and D.~Manocha, ``{Reciprocal n-body
  collision avoidance},'' in \emph{Proc. Int. Symp. Robot. Res. (ISRR)}, 2011,
  pp. 3--19.

\bibitem{vandenBerg2008}
J.~van~den Berg, M.~Lin, and D.~Manocha, ``{Reciprocal velocity obstacles for
  real-time multi-agent navigation},'' in \emph{Proc. IEEE Int. Conf. Robot.
  Autom. (ICRA)}, 2008, pp. 1928--1935.

\bibitem{yongjie2009collision}
Y.~Yongjie and Z.~Yan, ``Collision avoidance planning in multi-robot based on
  improved artificial potential field and rules,'' in \emph{Proc. IEEE Int.
  Conf. Robot. Biomimetics (ROBIO)}, 2009, pp. 1026--1031.

\bibitem{Zhou2017}
D.~Zhou, Z.~Wang, S.~Bandyopadhyay, and M.~Schwager, ``{Fast, on-line collision
  avoidance for dynamic vehicles using buffered voronoi cells},'' \emph{IEEE
  Robot. Autom. Lett.}, vol.~2, no.~2, pp. 1047--1054, 2017.

\bibitem{Zhu2019MRS}
H.~Zhu and J.~Alonso-Mora, ``B-uavc: buffered uncertainty-aware voronoi cells
  for probabilistic multi-robot collision avoidance,'' in \emph{Proc. Int.
  Symp. Multi. Robot. Multi. Agent. Syst. (MRS)}, 2019, pp. 162--168.

\bibitem{Wang2017}
L.~Wang, A.~D. Ames, and M.~Egerstedt, ``{Safety barrier certificates for
  collisions-free multirobot systems},'' \emph{IEEE Trans. Robot.}, vol.~33,
  no.~3, pp. 661--674, 2017.

\bibitem{Shi2020}
G.~Shi, W.~Honig, Y.~Yue, and S.-J. Chung, ``{Neural-swarm: decentralized
  close-proximity multirotor control using learned interactions},'' in
  \emph{Proc. IEEE Int. Conf. Robot. Autom. (ICRA)}, 2020, pp. 3241--3247.

\bibitem{Riviere2020}
B.~Riviere, W.~Honig, Y.~Yue, and S.~J. Chung, ``{Glas: global-to-local safe
  autonomy synthesis for multi-robot motion planning with end-to-end
  learning},'' \emph{IEEE Robot. Autom. Lett.}, vol.~5, no.~3, pp. 4249--4256,
  2020.

\bibitem{Chen2017a}
Y.~F. Chen, M.~Liu, M.~Everett, and J.~P. How, ``{Decentralized
  non-communicating multiagent collision avoidance with deep reinforcement
  learning},'' in \emph{Proc. IEEE Int. Conf. Robot. Autom. (ICRA)}, 2017, pp.
  285--292.

\bibitem{semnani2020multi}
S.~H. Semnani, H.~Liu, M.~Everett, A.~de~Ruiter, and J.~P. How, ``Multi-agent
  motion planning for dense and dynamic environments via deep reinforcement
  learning,'' \emph{IEEE Robot. Autom. Lett.}, vol.~5, no.~2, pp. 3221--3226,
  2020.

\bibitem{Luis2020}
C.~E. Luis, M.~Vukosavljev, and A.~P. Schoellig, ``{Online trajectory
  generation with distributed model predictive control for multi-robot motion
  planning},'' \emph{IEEE Robot. Autom. Lett.}, vol.~5, no.~2, pp. 604--611,
  2020.

\bibitem{Serra2020}
A.~Serra-G\'{o}mez, B.~Brito, H.~Zhu, J.~J. Chung, and J.~Alonso-mora, ``{With
  whom to communicate: learning efficient communication for multi-robot
  collision avoidance},'' in \emph{Proc. IEEE/RSJ Int. Conf. Intell. Robot.
  Syst. (IROS)}, 2020, pp. 11\,770--11\,776.

\bibitem{Schmerling2018}
E.~Schmerling, K.~Leung, W.~Vollprecht, and M.~Pavone, ``{Multimodal
  probabilistic model-based planning for human-robot interaction},'' in
  \emph{Proc. IEEE Int. Conf. Robot. Autom. (ICRA)}, 2018, pp. 3399--3406.

\bibitem{Fridovich-Keil2020}
D.~Fridovich-Keil, A.~Bajcsy, J.~F. Fisac, S.~L. Herbert, S.~Wang, A.~D.
  Dragan, and C.~J. Tomlin, ``{Confidence-aware motion prediction for real-time
  collision avoidance 1},'' \emph{I. J. Robot. Res.}, vol.~39, no. 2-3, pp.
  250--265, 2020.

\bibitem{Rudenko2020}
A.~Rudenko, L.~Palmieri, M.~Herman, K.~M. Kitani, D.~M. Gavrila, and K.~O.
  Arras, ``{Human motion trajectory prediction: a survey},'' \emph{I. J. Robot.
  Res.}, vol.~39, no.~8, pp. 895--935, 2020.

\bibitem{Helbing1995}
D.~Helbing and P.~Moln{\'{a}}r, ``{Social force model for pedestrian
  dynamics},'' \emph{Phys. Rev. E}, vol.~51, no.~5, pp. 4282--4286, 1995.

\bibitem{helbing1998generalized}
D.~Helbing and B.~Tilch, ``Generalized force model of traffic dynamics,''
  \emph{Phys. Rev. E}, vol.~58, no.~1, p. 133, 1998.

\bibitem{Turnwald2016}
A.~Turnwald, D.~Althoff, D.~Wollherr, and M.~Buss, ``{Understanding human
  avoidance behavior: interaction-aware decision making based on game
  theory},'' \emph{Int. J. Soc. Robot.}, vol.~8, no.~2, pp. 331--351, 2016.

\bibitem{Oyler2016}
D.~W. Oyler, Y.~Yildiz, A.~R. Girard, N.~I. Li, and I.~V. Kolmanovsky, ``{A
  game theoretical model of traffic with multiple interacting drivers for use
  in autonomous vehicle development},'' in \emph{Proc. Am. Control Conf.
  (ACC)}, 2016, pp. 1705--1710.

\bibitem{Kuderer2015}
M.~Kuderer, S.~Gulati, and W.~Burgard, ``{Learning driving styles for
  autonomous vehicles from demonstration},'' in \emph{Proc. IEEE Int. Conf.
  Robot. Autom. (ICRA)}, 2015, pp. 2641--2646.

\bibitem{Alahi2016}
A.~Alahi, K.~Goel, V.~Ramanathan, A.~Robicquet, L.~Fei-Fei, and S.~Savarese,
  ``{Social lstm: human trajectory prediction in crowded spaces},'' in
  \emph{Proc. IEEE Conf. Comput. Vis. Pattern Recognition (CVPR)}, 2016, pp.
  961--971.

\bibitem{Brito2020}
B.~Brito, H.~Zhu, W.~Pan, and J.~Alonso-mora, ``{Social-vrnn: one-shot
  multi-modal trajectory prediction for interacting pedestrians},'' in
  \emph{Proc. Conf. Robot Learn. (CoRL)}, 2020.

\bibitem{Lee2017}
N.~Lee, W.~Choi, P.~Vernaza, C.~B. Choy, P.~H. Torr, and M.~Chandraker,
  ``{Desire: distant future prediction in dynamic scenes with interacting
  agents},'' in \emph{Proc. IEEE Conf. Comput. Vis. Pattern Recognition
  (CVPR)}, 2017, pp. 2165--2174.

\bibitem{Gupta2018}
A.~Gupta, J.~Johnson, L.~Fei-Fei, S.~Savarese, and A.~Alahi, ``{Social gan:
  socially acceptable trajectories with generative adversarial networks},'' in
  \emph{Proc. IEEE Conf. Comput. Vis. Pattern Recognition (CVPR)}, 2018, pp.
  2255--2264.

\bibitem{mainprice2013human}
J.~Mainprice and D.~Berenson, ``Human-robot collaborative manipulation planning
  using early prediction of human motion,'' in \emph{Proc. IEEE/RSJ Int. Conf.
  Intell. Robot. Syst. (IROS)}, 2013, pp. 299--306.

\bibitem{park2019planner}
J.~S. Park, C.~Park, and D.~Manocha, ``I-planner: Intention-aware motion
  planning using learning-based human motion prediction,'' \emph{I. J. Robot.
  Res.}, vol.~38, no.~1, pp. 23--39, 2019.

\bibitem{Pfeiffer2018}
M.~Pfeiffer, G.~Paolo, H.~Sommer, J.~Nieto, R.~Siegwart, and C.~Cadena, ``{A
  data-driven model for interaction-aware pedestrian motion prediction in
  object cluttered environments},'' in \emph{Proc. IEEE Int. Conf. Robot.
  Autom. (ICRA)}, 2018, pp. 5921--5928.

\bibitem{Honig2018}
W.~Honig, J.~A. Preiss, T.~K. Kumar, G.~S. Sukhatme, and N.~Ayanian,
  ``{Trajectory planning for quadrotor swarms},'' \emph{IEEE Trans. Robot.},
  vol.~34, no.~4, pp. 856--869, 2018.

\bibitem{Scholler2020}
C.~Scholler, V.~Aravantinos, F.~Lay, and A.~Knoll, ``{What the constant
  velocity model can teach us about pedestrian motion prediction},'' \emph{IEEE
  Robot. Autom. Lett.}, vol.~5, no.~2, pp. 1696--1703, 2020.

\bibitem{Hochreiter1997}
S.~Hochreiter and J.~Urgen~Schmidhuber, ``{Long short-term memory},''
  \emph{Neural Comput.}, vol.~9, no.~8, pp. 1735--1780, 1997.

\bibitem{werbos1990backpropagation}
P.~J. Werbos, ``Backpropagation through time: what it does and how to do it,''
  \emph{Proc. IEEE}, vol.~78, no.~10, pp. 1550--1560, 1990.

\bibitem{domahidi2014forces}
A.~Domahidi and J.~Jerez, ``Forces professional,'' \emph{embotech GmbH
  (http://embotech.com/FORCES-Pro)}, 2014.

\end{thebibliography}

\end{document}